\documentclass{article}
\usepackage{animate}
\usepackage{listings}
\usepackage{xcolor}
\usepackage{caption}

\usepackage{arxiv}

\usepackage{algpseudocode}
\usepackage{natbib}
\usepackage{float}
\usepackage[linesnumbered,ruled]{algorithm2e}

\usepackage[utf8]{inputenc} 
\usepackage[T1]{fontenc}    
\usepackage{url}            
\usepackage{booktabs}  
\usepackage{blindtext}
\usepackage{amsfonts}       
\usepackage{nicefrac}       
\usepackage{microtype}      
\usepackage{xcolor}         

\usepackage{lipsum}                     %
\usepackage{xargs}
\usepackage{xfrac}
\usepackage{amsmath}
\usepackage{mleftright}
\usepackage{refcount} 

\usepackage{mathtools}   
\usepackage{amssymb}
\usepackage{enumitem}
\usepackage{amsthm}      
\usepackage{thmtools}    
\usepackage{mdframed}    

\newtheorem{theorem}{Theorem}
\newtheorem{lemma}[theorem]{Lemma}
\newtheorem{proposition}[theorem]{Proposition}
\newtheorem{corollary}[theorem]{Corollary}

\usepackage{lipsum}
\usepackage{xargs}
\usepackage{xfrac}
\usepackage{mleftright}

\usepackage{hyperref}
\usepackage{cleveref}    

\crefname{theorem}    {theorem}    {theorems}
\Crefname{theorem}    {Theorem}    {Theorems}

\crefname{lemma}      {lemma}      {lemmas}
\Crefname{lemma}      {Lemma}      {Lemmas}

\crefname{proposition}{proposition}{propositions}
\Crefname{proposition}{Proposition}{Propositions}

\crefname{corollary}  {corollary}  {corollaries}
\Crefname{corollary}  {Corollary}  {Corollaries}

\crefname{assumption}{assumption}{assumptions}
\Crefname{assumption}{Assumption}{Assumptions}

\renewcommand{\rm}{{\mathcal R}}

\usepackage{amsmath,amssymb}

\newcommand{\R}{{\mathbb{R}}}

\NewDocumentCommand{\numberthis}{om}{%
  \IfNoValueTF{#1}{%
    \refstepcounter{equation}\tag{\theequation}%
  }{%
    \tag{#1}%
  }%
  \label{#2}%
}

\def\[#1\]{\begin{align*}#1\end{align*}}

\usepackage{eqparbox}
\newcommand{\comment}[1]{}

\newcommand{\mmd}{\textrm{MMD}}

\usepackage{array}
    \newcolumntype{P}[1]{>{\centering\arraybackslash}p{#1}}
    \newcolumntype{M}[1]{>{\centering\arraybackslash}m{#1}}

\newcounter{savetheorem}

\newenvironment{repeatthm}[1]{%
  \setcounter{savetheorem}{\value{theorem}}%
  \edef\orignum{\getrefnumber{#1}}%
  \setcounter{theorem}{\numexpr\orignum-1\relax}%
  \begin{theorem}%
}{%
  \end{theorem}%
  \setcounter{theorem}{\value{savetheorem}}%
}

\newenvironment{repeatlemma}[1]{%
  \setcounter{savetheorem}{\value{theorem}}%
  \edef\orignum{\getrefnumber{#1}}%
  \setcounter{theorem}{\numexpr\orignum-1\relax}%
  \begin{lemma}%
}{%
  \end{lemma}%
  \setcounter{theorem}{\value{savetheorem}}%
}

\newenvironment{repeatprop}[1]{%
  \setcounter{savetheorem}{\value{theorem}}%
  \edef\orignum{\getrefnumber{#1}}%
  \setcounter{theorem}{\numexpr\orignum-1\relax}%
  \begin{proposition}%
}{%
  \end{proposition}%
  \setcounter{theorem}{\value{savetheorem}}%
}

\AtEndPreamble{
\hypersetup{
    colorlinks=true,       
    linkcolor=blue,        
    citecolor=blue,         
    filecolor=magenta,     
    urlcolor=cyan          
}
}

\title{Data Generation without Function Estimation}
\usepackage{authblk}
\usepackage{blindtext}
\begin{document}

\author[1]{Hadi Daneshmand}
\author[2]{Ashkan Soleymani}
\affil[1]{Department of Computer Science, University of Virginia }
\affil[2]{Department of Electrical Engineering and Computer Science, MIT}
\affil[]{\textit {dhadi@virginia.edu}}
\maketitle

\begin{abstract}
  Estimating the score function—or other population-density-dependent functions —is a fundamental component of most generative models. However, such function estimation is computationally and statistically challenging. \emph{Can we avoid function estimation for data generation?} We propose an \textbf{estimation-free} generative method: \emph{A set of points} whose locations are \emph{deterministically} updated with (inverse) \emph{gradient descent} can transport a uniform distribution to arbitrary data distribution, in the mean field regime, \textbf{without function estimation, training neural networks, and even noise injection}. The proposed method is built upon recent advances in the physics of interacting particles. We show, both theoretically and experimentally, that these advances can be leveraged to develop novel generative methods.   
\end{abstract}

\section{Introduction}
Given i.i.d. samples from an unknown distribution, how to generate a new sample from the distribution?
Existing generative models often rely on estimating functions depending on population density such as the score function. Such an estimation is both statistically and computationally challenging. Computationally, estimating the score function even for a simple Gaussian mixture model with maximum likelihood estimation of parameters is NP-hard~\citep{arora2001learning}. Statistically, score estimation suffers the curse of dimensionality~\citep{wibisono2024optimal}. This raises a fundamental question: can we avoid function estimation for data generation? Ultimately, what is typically available during training is an empirical distribution over i.i.d. samples, for which the score function is not even defined. Can we generate new samples directly from this discrete empirical distribution, avoiding (score) function estimation?

We demonstrate that data generation can be achieved without function estimation. Our proposed approach is a novel generative method operating entirely on the empirical distribution of a finite set of training samples. We construct an interactive system that iteratively updates the positions of these data points in two phases: (1) applying standard gradient descent to shape data points (2) performing an inverse gradient descent step to generate new samples.  This method builds upon recent advances in the theoretical study of systems of interacting particles~\citep{duerinckx2020mean,frank2025minimizers}. Leveraging this rich literature, we prove that the proposed estimation-free method can transport a uniform measure over a finite ball/sphere to an arbitrary data distribution in the mean-field regime, where the number of samples tends to infinity. Finally, we experimentally validate estimation-free data generation using a finite number of points.

\section{Method}

\begin{figure}[t!]
    \centering
\begin{tabular}{cccc}    \includegraphics[width=0.23\textwidth]{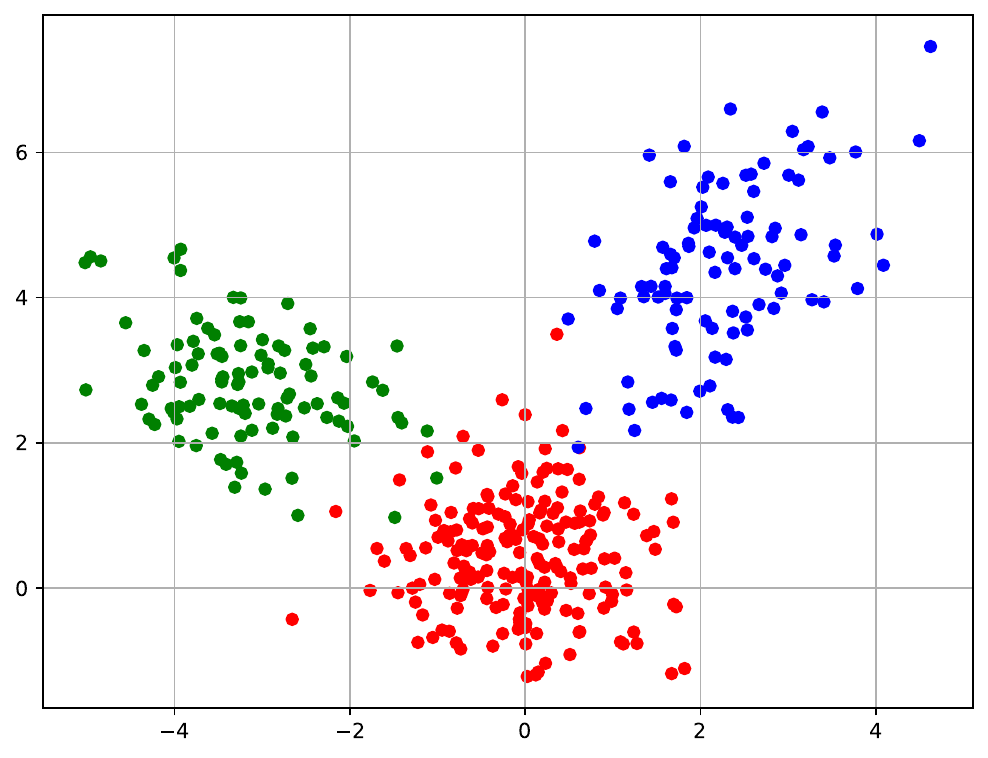} & \includegraphics[width=0.23\textwidth]{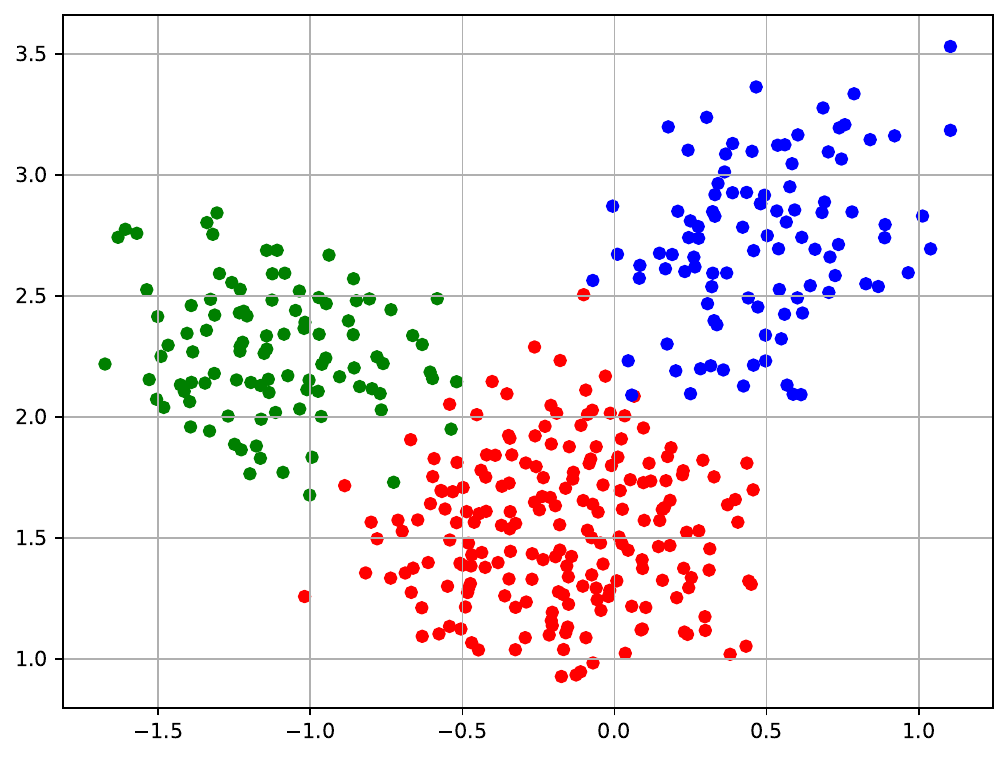}  & \includegraphics[width=0.23\textwidth]{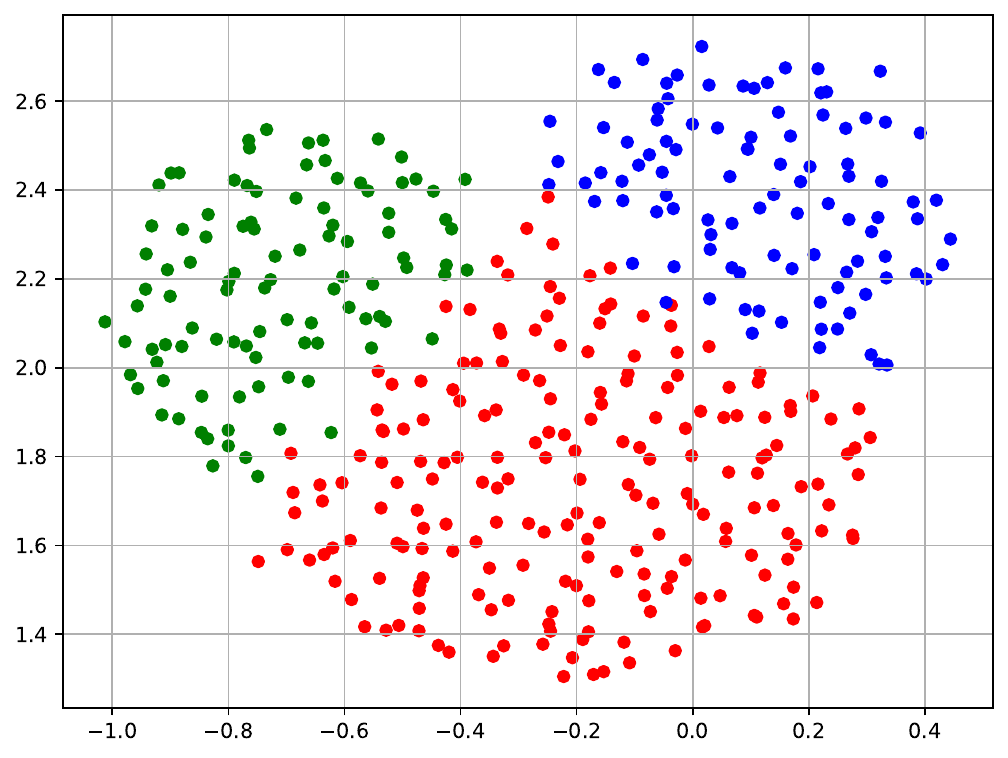} & \includegraphics[width=0.23\textwidth]{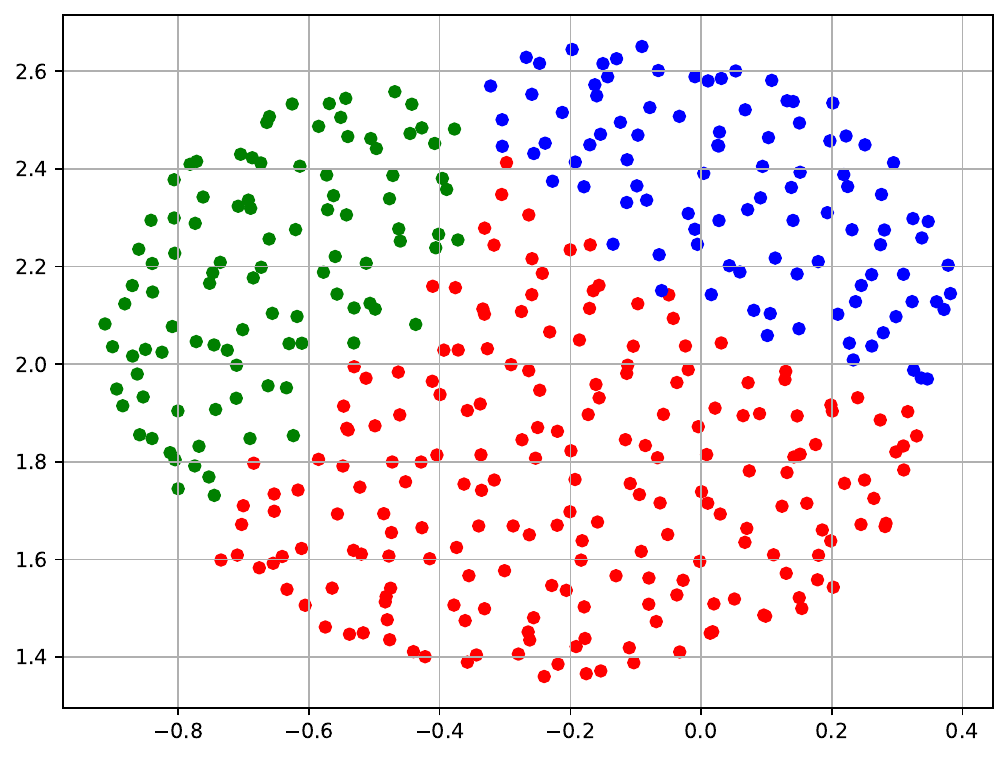}\\
    $\{x_i^{(0)}\}$ $\Rightarrow$ &  $\{x_i^{(6)}\}$ $\Rightarrow$ &  $\{x_i^{(12)}\}$ $\Rightarrow$ & $\{x_i^{(18)}\}$
\end{tabular}
    \caption{\footnotesize{\textbf{Convergence of the empirical distribution with optimization.} } Scatter plot of gradient descent iterates $x_i^{(k)}$, defined in \eqref{eq:gd}, initialized at $x_i^{(0)}$ drawn from a Gaussian mixture distribution. Different mixture components (modes) are distinguished by color. As $k$ evolves, the points become uniformly distributed on the circle. }
    \label{fig:forward}
\end{figure}

Given the support of an empirical distribution over points $x_1, \dots, x_n \in \mathbb{R}^d$, consider the following optimization problem:
\begin{align}
    x_1^*, \dots, x_n^* := \arg\min_{x_1, \dots, x_n \in \mathbb{R}^d} \left( E_n^{(\epsilon)}(x_1, \dots, x_n) := \frac{1}{n(n-1)} \sum_{i=1}^n \sum_{j \neq i} W_{\epsilon}^{(s)}(x_i - x_j) \right),
\end{align}
where 
$W^{(s)}_{\epsilon}(x - y) = \frac{\|x - y\|^2}{2} +  \frac{1}{s(\|x - y\|^2+\epsilon)^{s/2}}$
is called a "attractive-repulsive
power-law interaction potential"~\citep{balague2013nonlocal,shu2025family}. The squared-norm term encourages attraction between particles, while the inverse-norm term induces repulsion, preventing collapse\footnote{For \( s = 0 \), the repulsive term is replaced by the logarithm of the norm, as given by
 $s\to 0$~\citep{shu2025family}.}.  These competing forces lead to a striking structure in the distribution of the optimized points $x_1^*, \dots, x_n^*$ in the asymptotic regime as $n \to \infty$, where the limiting distribution is characterized by
\begin{align}
    \arg\min_{p \in \Omega} \left( E(p) := \int W^{(s)}(x - y) \, p(x) \, p(y) \, dx \, dy \right),  \Omega := \{ \text{probability measures over } \mathbb{R}^d \} \numberthis{eq:opt_mean_field},
\end{align}
 where $W^{(s)}:=W^{(s)}_0$.
The minimizer is unique up to translation and is uniform over 
\[  \begin{cases}
        \text{a ball  ~\citep{carrillo2023radial}} & \text{if } s=d-2 , \\
        \text{a sphere~\citep{frank2025minimizers2}} & \text{if } 2<d \text{ and } -2\leq s<d-4 ,
    \end{cases}
\]
with a radius that is finite and computable in closed form.
By optimizing the locations $x_1, \dots, x_n$, one can asymptotically transport any empirical distribution to a simple uniform distribution over a compact ball or sphere. This optimization can be performed via standard gradient descent (GD):
\begin{align} \label{eq:gd}
    \forall i: \quad x_i^{(k+1)} = x_i^{(k)} - \gamma \nabla_{x_i} E_n^{(\epsilon)}(x_1^{(k)}, \dots, x_n^{(k)}).
\end{align}
Figure~\ref{fig:forward} plots the time evaluation of points $\{x_i^{(k)}\}$ indexed by $k$ starting from the initial points $x_i^{(0)}$ drawn i.i.d. from Gaussian mixture. We observe that the distribution becomes uniform as $k$ increases.
Although the distribution is uniform, it contains very interesting information about the initial Gaussian mixture distribution. For example, the ratio of area of different colors are proportional to the number of points in each cluster. This structure indicates that it may be possible to recover original data distribution from the final state. Whereas in the initial state \(x_i^{(0)}\) the data points are i.i.d., evolving according to \eqref{eq:gd} induces strong correlations and entanglement among them under the gradient descent dynamics. Thus, in the limit \(k \to \infty\), although the marginal distribution of each data point becomes uniform, their joint distribution still conveys potentially useful information about the original distribution.

Even more interesting is the inverse of gradient descent: assuming GD converges to the minimizer of $E_n$, inverse GD maps points uniformly distributed on a sphere (or disk) to an arbitrary empirical distribution supported on a finite set of points. The following proposition guarantees that this inverse exists and is computable under mild assumptions.

\begin{proposition} \label{prop:proximal}
    Consider the following proximal optimization problem 
    \begin{align} 
    (y^*_1, y^*_2, \cdots, y^*_n) = \arg\min_{y_1, y_2, \cdots, y_n \in \mathbb{R}^d} \frac{1}{2} \sum_{i = 1}^d \| y_i -x_i^{(k)} \|^2 - \gamma E^{(\epsilon)}_{n}(y_1, y_2, \cdots, y_{n}) \numberthis{eq:prox_inverse}
    \end{align}
    The above optimization is convex with solution $y_i^* = x_i^{(k-1)}$ for all $i \in [n]$, as long as the learning rate $\gamma$ is sufficiently small.
\end{proposition}
Thus, the inverse process can be efficiently computed using gradient descent on a convex function, without requiring the score function or any other statistical information about the population density. In fact, both gradient descent (GD) and its inverse can deterministically transport an empirical distribution to a uniform distribution and vice versa. This insight enables a three-step sampling method that operates on a finite set of points: (1) apply GD to the $n$ available samples from the target distribution; (2) add a new point drawn uniformly at random from a sphere or ball; and (3) apply inverse GD. We call this three step method \textbf{estimation-free sampling (EFS)}.

\emph{(1) Forward optimization:}  Apply gradient descent to $E_n^{(\epsilon)}$ (see~\eqref{eq:gd}) starting from i.i.d. samples $x_1^{(0)}, \dots, x_n^{(0)}$ drawn from the target distribution. In the next section, we show that this optimization acts as a transport map, pushing the empirical distribution toward a uniform distribution in the asymptotic regime.

\emph{(2) Augmentation:} Add a new point $y^{(k)}$, sampled from a uniform distribution, to the set ${x_0^{(k)}, \dots, x_n^{(k)}}$. To generate $y^{(k)}$, we first estimate the center and radius of the ball or sphere enclosing the points $x_i^{(k)}$. $y^{(k)}$ can also be generated by interpolating between two points in the set, i.e., ${y^{(k)} = (1 - t) x_i^{(k)} + t x_j^{(k)}}$ for some $i \neq j$ and $t \in (0,1)$.

\begin{figure}[H]
\begin{minipage}{0.43\textwidth}
\begin{algorithm}[H]
\caption{Backward Optimization} \label{alg:backward}
\KwIn{Data $\{x_1^{(j)},\dots, x_n^{(j)}\}_{j=1}^k$, new sample $y^{(k)}$}

\KwSty{}{\textbf{Parameters}: $\gamma$, $k$, $\beta$, $T$, $s$ and $\epsilon$}

\KwSty{}{\textbf{Set}: $j=k$ }

\While{$j\geq 0$}{
    \textbf{Set}: $v_0 = y^{(j)}$

    \For{$t \leftarrow 0$ \KwTo $T$}{
    $\nabla = \tfrac{\gamma}{n} \sum_{i \in [n]} \nabla W_\epsilon^{(s)}(v_t - x^{(j)}_i)$
    
    $\Delta = v_t - y^{(j)} - \nabla $
    
    $v_{t+1} = v_t - \beta  \Delta $
    }
    $j = j-1$ and $y^{(j)} = v_T$
    } 
\KwOut{$y^{(0)}$}
\end{algorithm}
\end{minipage}
\begin{minipage}{0.54\textwidth}

\begin{tabular}{cc}
\includegraphics[width=0.5\textwidth] {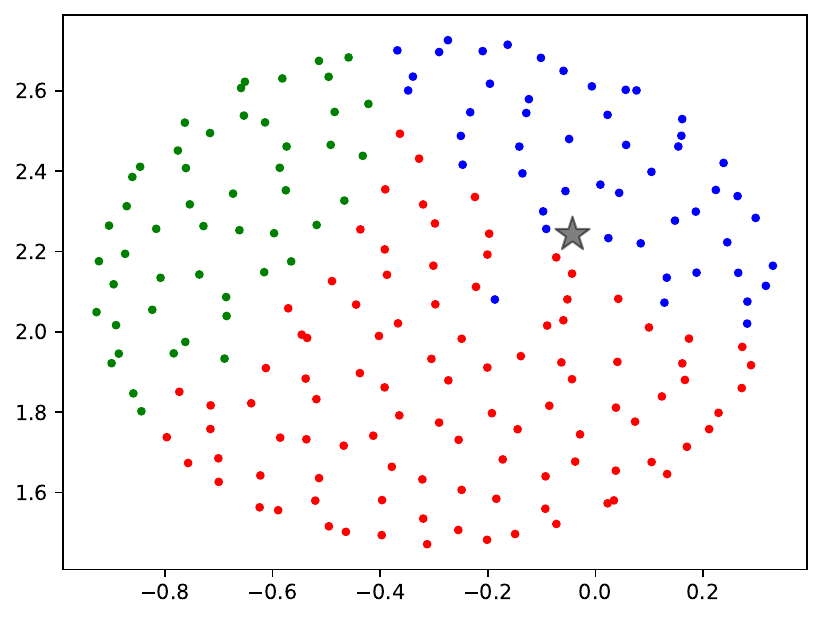} & \includegraphics[width=0.5\textwidth] {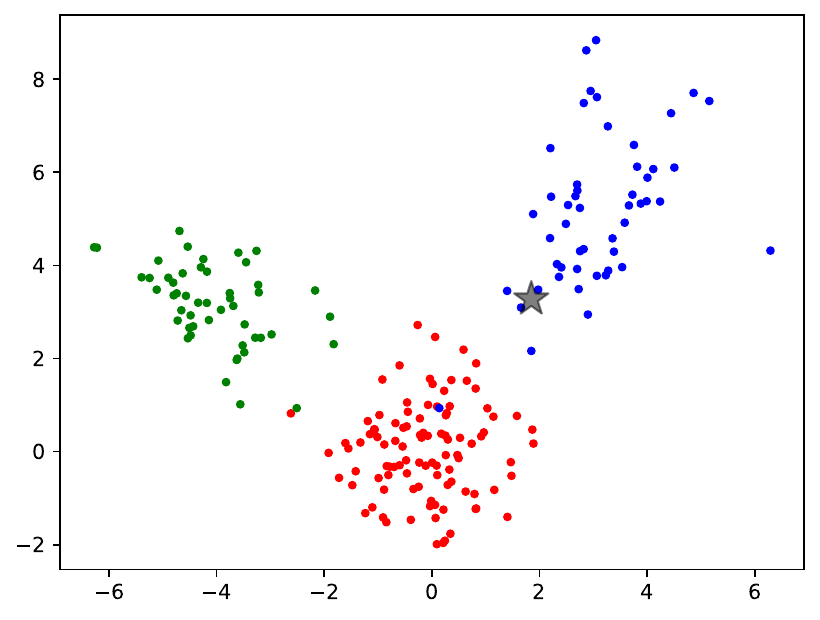} \\
  $\{x_i^{(k)}\} \cup \{ y^{(k)} =\star \}$   & $\{x_i^{(0)}\} \cup \{ y^{(0)} =\star\}$
\end{tabular}
\caption{\footnotesize{\textbf{Sampling:} $\star$: generated sample; Left: $x_i^{(0)} \stackrel{i.i.d.}{\sim}$ mixture Gaussian; Right: $x_i^{(k)}$ obtained by forward optimization. Colors: mixture components. The location of the new sample is updated by Backward Optimization~\ref{alg:backward}}} \label{fig:backward}
\end{minipage}
\end{figure}

\emph{(3) Backward optimization:}
Compute the inverse GD for the newly generated point $y^{(k)}$ by optimizing the convex function defined in Proposition~\ref{prop:proximal}, using efficient gradient descent with a constant step size. This backward optimization is detailed in Algorithm~\ref{alg:backward}. While the forward process involves $n$ points, the backward step involves $n+1$ points, consisting of the $n$ points from the forward step and the newly generated point from step (2).
Find Algorithm details in Appendix~\ref{section:algorithm}. Figure~\ref{fig:backward} illustrates the full sampling procedure from a Gaussian mixture using steps (1)–(3). Specifically, it shows the outcome of applying the backward optimization to a point initially drawn from a uniform distribution over a 2D circle. The transformed point is mapped into one of the modes of the Gaussian mixture. Interestingly, the initial location on the circle often predicts which mixture component the point will align with after optimization. Additional results are provided in Section~\ref{sec:experiments}.  
\section{Theory} \label{section:theory}
We demonstrate that the proposed estimation-free sampling (EFS) method can generate new samples from a distribution in the asymptotic regime. As established in Proposition~\ref{prop:proximal}, backward step is convex problem with well-established theoretical guarantees. In contrast, the forward optimization is non-convex. Nevertheless, its global convergence can be analyzed using recent advances in Wasserstein gradient flows on interactive-repulsive energy~\citep{duerinckx2020mean,carrillo2023radial}. As the potential function is shift-invariant, \emph{all statements hold up to translation}, and we refrain from repeating "up to translation" for simplicity.

\subsection{Forward Transport to Uniform Distribution} 
While our ultimate goal is to analyze the backward step of EFS, which generates new samples, the forward step also plays a critical role in data generation. Here, we show that the forward step of EFS \emph{transports} the data distribution to a uniform distribution\footnote{A map \( f \) transports measure \( \mu \) to \( \nu \) if the distribution of \( f(x) \) is \( \nu \) when \( x \sim \mu \)~\citep{villani2008optimal}.}.

Consider a continuous-time model where the samples \( x_i^{(k)} \) updated by gradient descent, modeled by the following ordinary differential equation (ODE):
\begin{align} 
\frac{dx_i}{dt} = - \frac{1}{n} \sum_{j \neq i} \nabla W^{(s)}(x_i - x_j), \label{eq:forwardode}
\end{align}
ODEs have been widely used to describe the limiting behavior of gradient descent with an infinitesimally small step size~\citep{su2016differential,zhang2021rethinking,zhang2021revisiting,chizat2018global}. Analyzing the above system becomes challenging due to the coupling between the variables \( x_i \), as the complexity increases with the number of particles \( n \). A common approach to address this challenge is to analyze the evolution of the empirical distribution of the particles at a macroscopic level, rather than tracking their individual trajectories. Define the empirical distribution as
\begin{align}\label{eq:empirical}
\mu_t^{(n)} := \frac{1}{n} \sum_{i=1}^n \delta_{x_i(t)},
\end{align}
where \( \delta_x \) denotes the Dirac measure centered at \( x \). \citet{duerinckx2020mean} prove that, in the limit \( n \to \infty \), the empirical measure \( \mu_t^{(n)} \) converges to a solution of a gradient flow on the space of probability measures, optimizing an energy functional \( E \) with respect to the Wasserstein-2 metric. This type of abstract optimization generalizes Euclidean gradient descent to general metric spaces~\citep{ambrosio2005gradient} as 
\begin{align}
\mu^{(\gamma)}_{k+1} = \arg\min_{\mu \in \Omega} \frac{1}{2} \mathcal{W}_2^2(\mu_k^{(\gamma)}, \mu) + \gamma E(\mu),
\end{align}
where \( \mathcal{W}_2(\cdot, \cdot) \) denotes the Wasserstein-2 distance. A continuous-time interpolation of the discrete sequence \( \mu_k^{(\gamma)} \) converges to a measure \( \mu_t \) as \( \gamma \to 0 \), which satisfies the following continuity equation:
\begin{align}
\frac{d\mu}{dt} = \mathrm{div}\left( \mu(x) \int \nabla W^{(s)}(x - y) \mu(y) \, dy \right), \label{eq:continuity}
\end{align}
where \( \mathrm{div} \) denotes the divergence operator acting on the associated vector field. \citet{duerinckx2020mean} prove \( \mu_t^{(n)} \) converges in the weak sense to \( \mu_t \), the solution of the PDE above.

\begin{theorem}[\citet{duerinckx2020mean}] \label{thm:weak_convergenc}
If \( \mu_0^{(n)} \) converges to a regular measure \( \mu_0 \) in Wasserstein distance, then \( \mu_t^{(n)} \) converges weakly to \( \mu_t \), provided \( d - 2 \leq s < d \).
\end{theorem}

The macroscopic measure \( \mu_t \) is significantly easier to analyze than the microscopic trajectories \( x_i(t) \). While the energy \( E_n \) is generally non-convex in \( x_i(t) \), the limiting energy \( E \) obeys linear interpolation convexity in \( \mu \)~\citep{shu2025family}. This convexity can be exploited to characterize the asymptotic behavior of \( \mu_t \) as \( t \to \infty \)~\citep{shu2025family}.

\begin{theorem}[\cite{frank2025minimizers, carrillo2023radial}] \label{thm:frank}
The steady state of \( \mu_t \) is the global minimizer of \( E \) (up to translation) which is the uniform measure over:
\[
\begin{cases}
\text{a ball~\citep{carrillo2023radial}} & \text{if } d - 2 \leq s < d, \\
\text{a sphere~\citep{frank2025minimizers2}} & \text{if } 2 < d \text{ and } -2 \leq s < d - 4,
\end{cases}
\]
with finite radius.
\end{theorem}

Notably, proving that a steady state (i.e., a local minimizer) is in fact a global minimizer required decades of mathematical development, beginning with the foundational work of \citet{frostman1935potentiel}, and has only recently been fully resolved in certain parameter regimes~\citep{frank2025minimizers}. These advances allow us to analyze the asymptotic dynamics of the forward step in EFS.

\begin{corollary} \label{cor:largem}
If \( \mu_0^{(n)} \) converges to \( \mu_0 \) in Wasserstein-2 distance, then the measure \( \mu_t^{(n)} \) converges in the weak sense to the uniform distribution over a ball of finite radius as \( n \to \infty \) and \( t \to \infty \), for \( s = d - 2 \).
\end{corollary}

\subsection{Backward Transport to Target Distribution}
It is straightforward to verify that the backward optimization can exactly recover the original training data $x_k^{(i)}$ by initializing Algorithm~\ref{alg:backward} with $y_k = x_k^{(i)}$. However, our goal is not to reconstruct existing data points, but to generate new samples from the underlying data distribution. We show that this form of generalization is achievable in the asymptotic regime.

Consider the continuous-time formulation of the backward optimization process:
\begin{align}
    \frac{dy^{(n)}}{d\bar{t}} = \frac{1}{n} \sum_{i=1}^n \nabla W^{(s)}\left(y^{(n)}_t - x_i(t)\right), \quad \text{where } x_i(t) \text{ is defined in~\eqref{eq:forwardode}}. \label{eq:backwardode}
\end{align}
Here, the differential $d\bar{t}$ indicates that the process is reversed time~\citep{anderson1982reverse}. This dynamics corresponds to Algorithm~\ref{alg:backward} in the limit of an infinitesimally small step size $\gamma$ and a large terminal time $T \to \infty$. We prove this dynamics \emph{transport}  $\mu_t$ back to $\mu_0$ in the asymptotic regime as $n \to \infty$. In other words, the distribution of $y_0^{(n)}$ converges to $\mu_0$ as $n\to \infty$.

\begin{theorem} \label{thm:backward}
Assume $\mu_0^{(n)}$ converges to $\mu_0$ in Wasserstein-2 distance. 
    Suppose that \( y_t^{(n)} \) is a random variable with law \( \mu_t \), where \( \mu_t \) is the solution to the continuity equation~\eqref{eq:continuity}. Then, the distribution of \( y_0^{(n)} \)—obtained from the reversed-time ODE~\eqref{eq:backwardode}—converges to \( \mu_0 \) as \( n \to \infty \), for continuous measures \( \mu_0 \) and \( \mu_t \), and for \( s=d-2 \).
\end{theorem}

To generate new samples, we assume access to i.i.d. samples from the target distribution. Since the empirical measure \( \mu_0^{(n)} \) converges to \( \mu_0 \) in Wasserstein distance~\citep{villani2008optimal} which ensures the assumption holds in the last theorem. According to Corollary~\ref{cor:largem}, the measure \( \mu_t \) converges to a uniform distribution as $t\to \infty$. Consequently, sampling from a nearly uniform distribution allows for effective recovery of the initial data distribution \( \mu_0 \).

\begin{figure}[t!]
\centering
   \begin{tabular}{c c}
   Original Samples & Transported Samples \\
\includegraphics[width=0.4\textwidth]{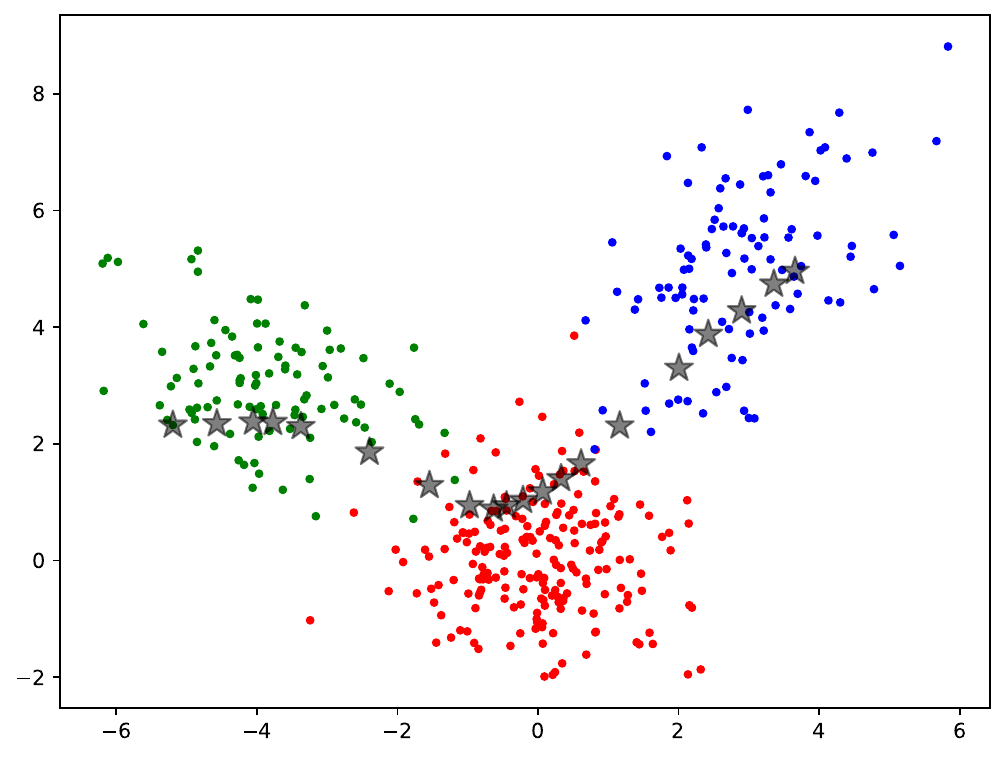}   & \includegraphics[width=0.4\textwidth]{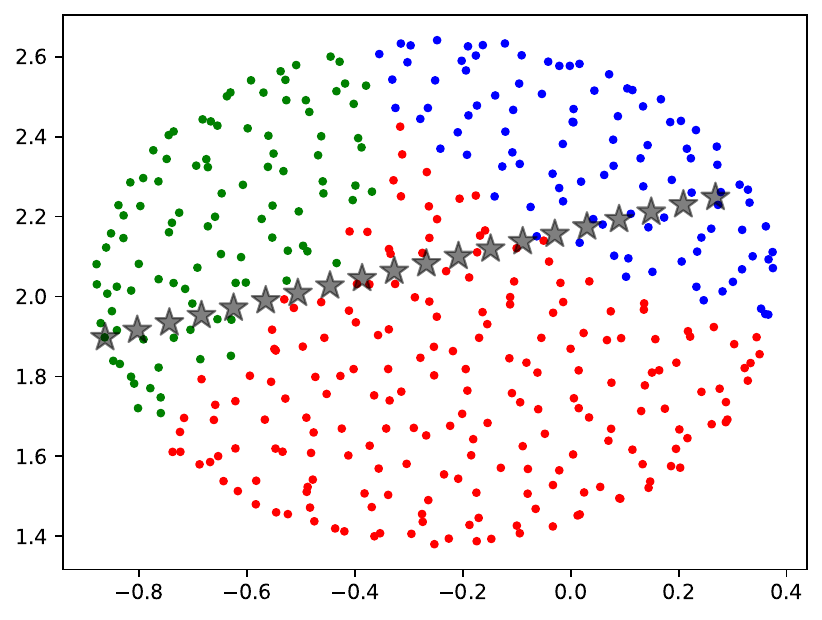} \\
        & 
   \end{tabular}
    \caption{\footnotesize{\textbf{Interpolation in Latent Space.} \emph{Colored points} are samples drawn from a Gaussian mixture model (left) and their positions after forward optimization (right). The $\star$s are the generated sample $y$ in EFS (right), and their location after applying  Algorithm~\ref{alg:backward} (left). The straight line in the latent space (right) is transformed into a curved trajectory (left) that aligns with the data distribution. Colors correspond to different mixture components.}}
    \label{fig:mixture_interpolation}
\end{figure}
Thus, \textbf{a set of points} whose locations are updated via \emph{(inverse) gradient descent}, can provably generate new samples from an arbitrary distribution $\mu_0$—without function estimation, training neural networks, or injecting noise. 

Remarkably, diffusion models use the Kolmogorov backward equation to transport a Gaussian distribution to an arbitrary target distribution, namely the following reverse-time stochastic differential equation~\citep{anderson1982reverse}:
\begin{align}
\frac{dy}{d\bar{t}} = (-x_t - \nabla_x p_t(x)) d\bar{t} + dW_{\bar{t}},
\end{align}
where $W_{\bar{t}}$ denotes reverse-time Brownian motion. The above SDE requires access to the score function $\nabla p_t(x)$, where $p_t$ is the population density at time $t$ associated with the diffusion process. Even when $p_t$ is a Gaussian mixture, estimating the score is challenging, as maximum likelihood estimation of its parameters is a non-convex problem~\citep{arora2001learning}. In contrast, our proposed model can efficiently sample from Gaussian mixtures, as illustrated in Figure~\ref{fig:backward}.

\section{Experiments} \label{sec:experiments}

While we have proven EFS can transport an almost uniform distribution to an arbitrary target distribution, our result is asymptotic—it holds in the limit as \( n \to \infty \). This asymptotic regime greatly simplifies the theoretical analysis. However, practical applications are inherently constrained to finite \( n \). We bridge this gap with experiments.

As the first estimation-free generative algorithm, EFS needs further research for a comprehensive benchmarking. Our experiments underscore both the practical challenges and the promising potential of this novel  generative method. The code is implemented in PyTorch~\citep{paszke2019pytorch} and was executed on a single NVIDIA RTX 6000 GPU with 48GB memory. The total execution time is under 15 minutes. 

\paragraph{Mixture of Gaussians: Capturing Data Geometry.}  
EFS effectively recovers the underlying data distribution even after transforming  into a set of points uniformly distributed on a circle. Figure~\ref{fig:mixture_interpolation} displays the original samples drawn from a Gaussian mixture (left) and their transformed positions after the EFS forward optimization step (right), where they are mapped onto the circle. New samples are then generated along a straight line on circle. Backward optimization maps these linear samples onto a curved trajectory that closely follows the geometry of the original data distribution. The backward process adaptively distorts pairwise distances to avoid low-density regions. Furthermore, we observe that EFS creatively generate unseen new samples.

\paragraph{Generating  samples from Swiss roll.}
We repeat the experiments on the Swiss roll dataset, similar to those on the Gaussian mixture in Figures~\ref{fig:forward}, \ref{fig:backward}, and \ref{fig:mixture_interpolation}. Table~\ref{table:params} summarizes the parameter choices for this dataset. We evaluate: (i) convergence to a uniform distribution via forward optimization, (ii) generation of new samples via backward optimization, and (iii) preservation of data geometry through interpolation.
 
\begin{figure}[t!]
    \centering
\begin{tabular}{cccc}    \includegraphics[width=0.24\textwidth]{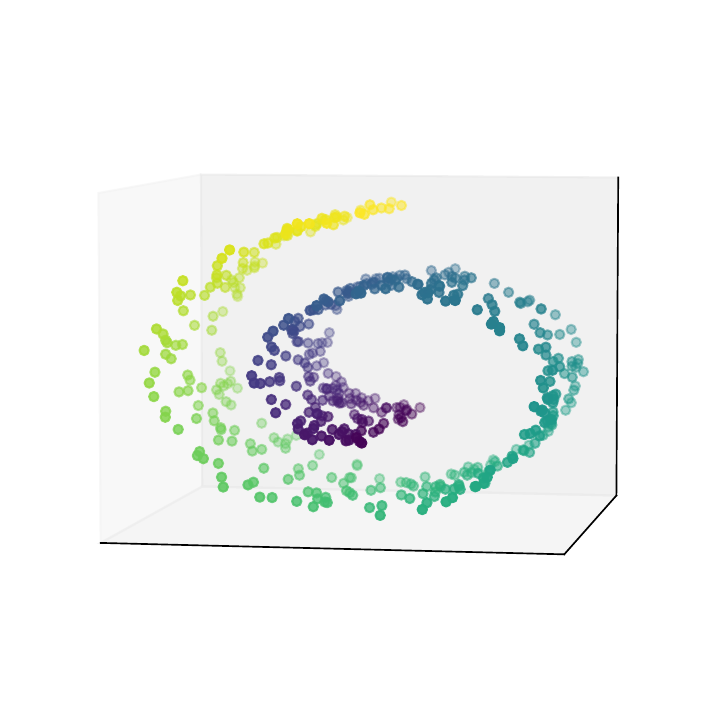} & \includegraphics[width=0.24\textwidth]{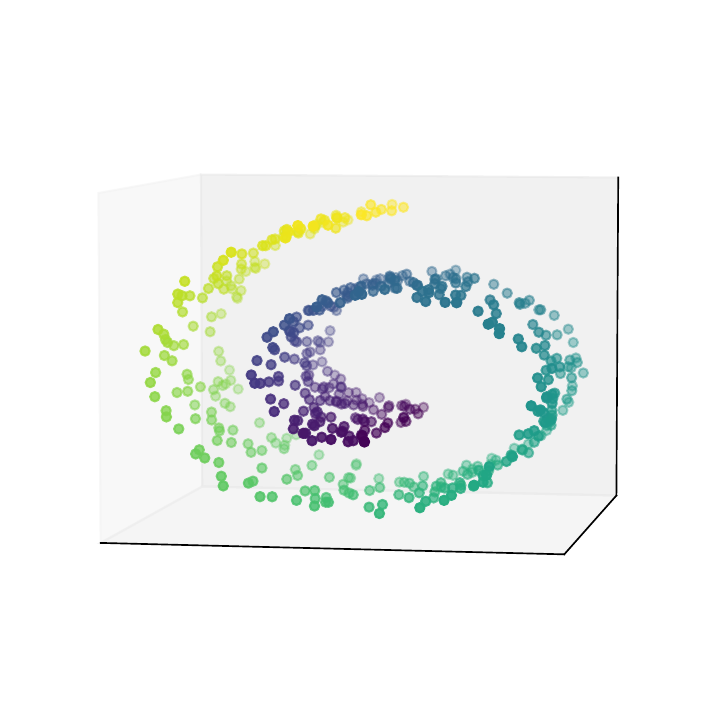}  & \includegraphics[width=0.24\textwidth]{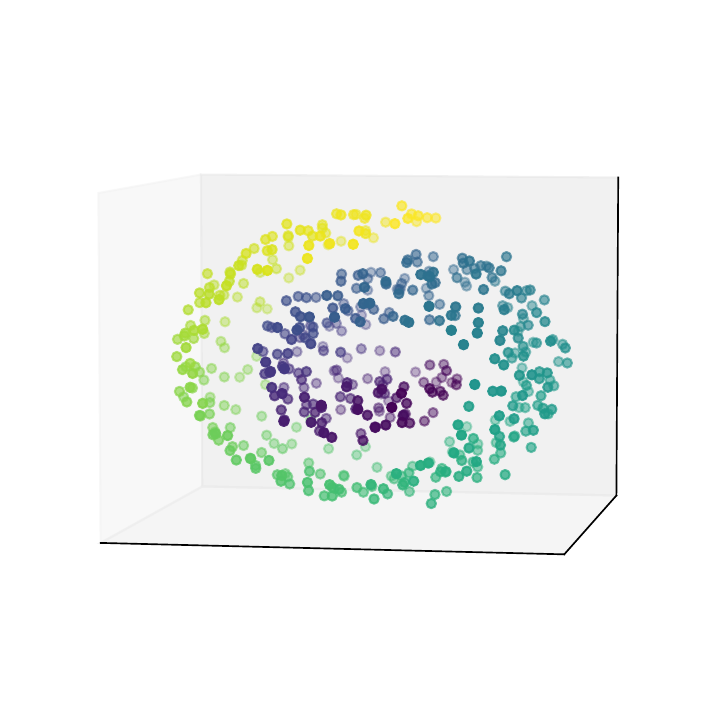} & \includegraphics[width=0.24\textwidth]{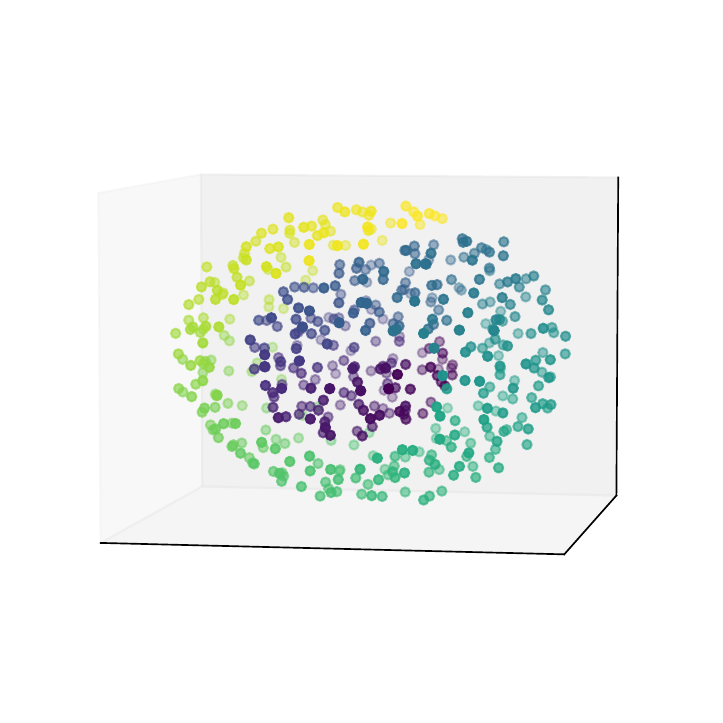}\\
    $\{x_i^{(0)}\}$ $\Rightarrow$ &  $\{x_i^{(36)}\}$ $\Rightarrow$ &  $\{x_i^{(72)}\}$ $\Rightarrow$ & $\{x_i^{(108)}\}$
\end{tabular}
    \caption{\footnotesize{\textbf{Forward convergence for Swiss roll.} } Scatter plot of gradient descent iterates $x_i^{(k)}$, defined in \eqref{eq:gd}, initialized at $x_i^{(0)}$ drawn from a Swiss roll with noise level 0.2. As $k$ evolves, the points become uniformly distributed akin to 2d example of Gaussian mixture in Figure~\ref{fig:forward}. Notably, Corollary~\ref{cor:largem} establishes this convergence in the mean-field regime as \( n \to \infty \).
 }
    \label{fig:forward_appendix}
\end{figure}

\begin{figure}
\centering
   \begin{tabular}{c c}
\includegraphics[width=0.3\textwidth] {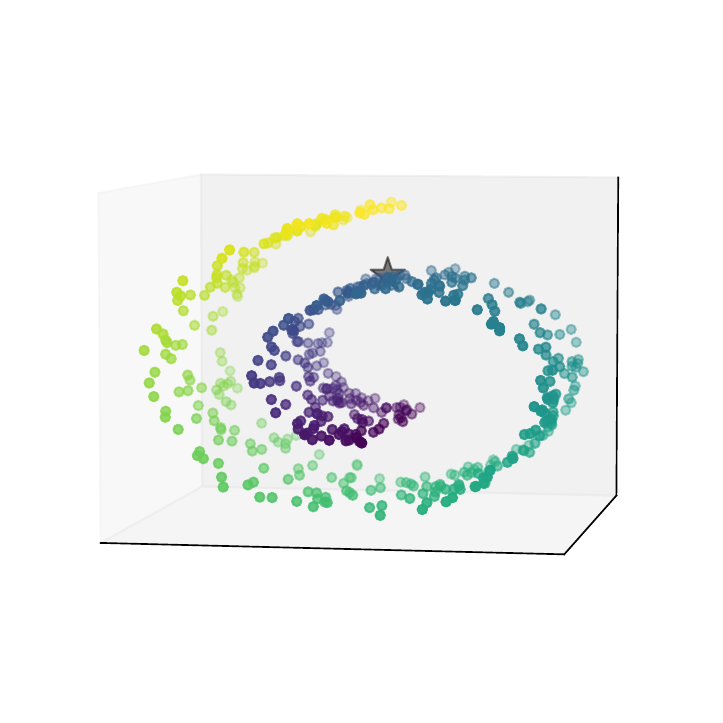} & \includegraphics[width=0.3\textwidth] {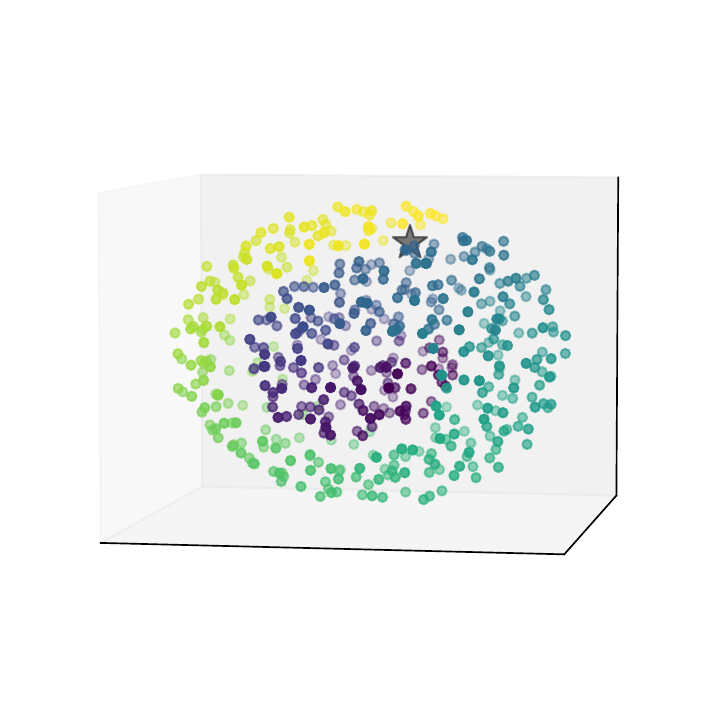} \\
   $\{x_i^{(0)}\} \cup \{ y^{(0)} =\star\}$ & $\{x_i^{(k)}\} \cup \{ y^{(k)} =\star \}$  
\end{tabular}
\caption{\footnotesize{\textbf{Sampling:} The black $\star$ is the generated sample $y^{(0)}$ ; Left: $x_i^{(0)} \stackrel{i.i.d.}{\sim}$ Swiss roll with noise level $0.2$; Right: $x_i^{(k)}$ obtained by forward optimization. Colors: mixture components. The location of the new sample is updated by Backward Optimization~\ref{alg:backward}}} \label{fig:backward_appendix}
\end{figure}

\begin{figure}[t!]
\centering
   \begin{tabular}{c c}
\includegraphics[width=0.35\textwidth]{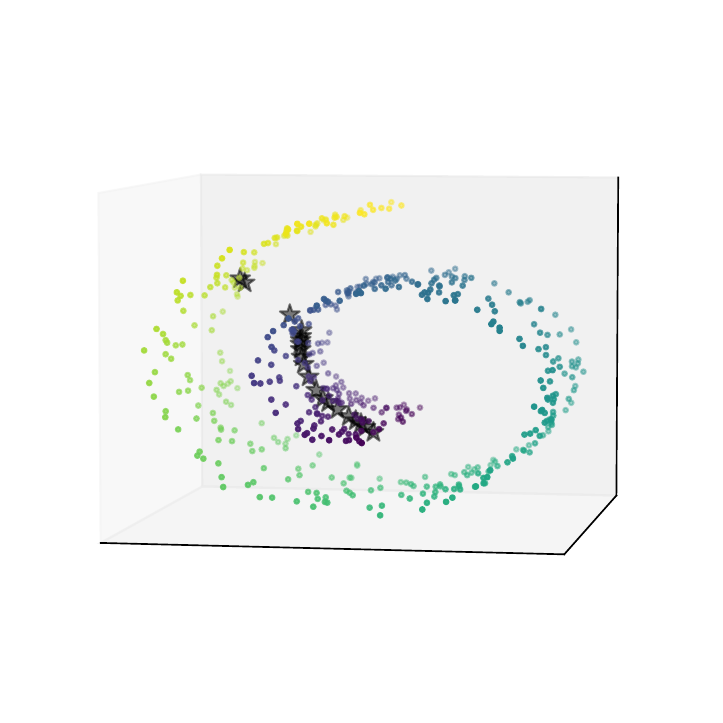}   & \includegraphics[width=0.35\textwidth]{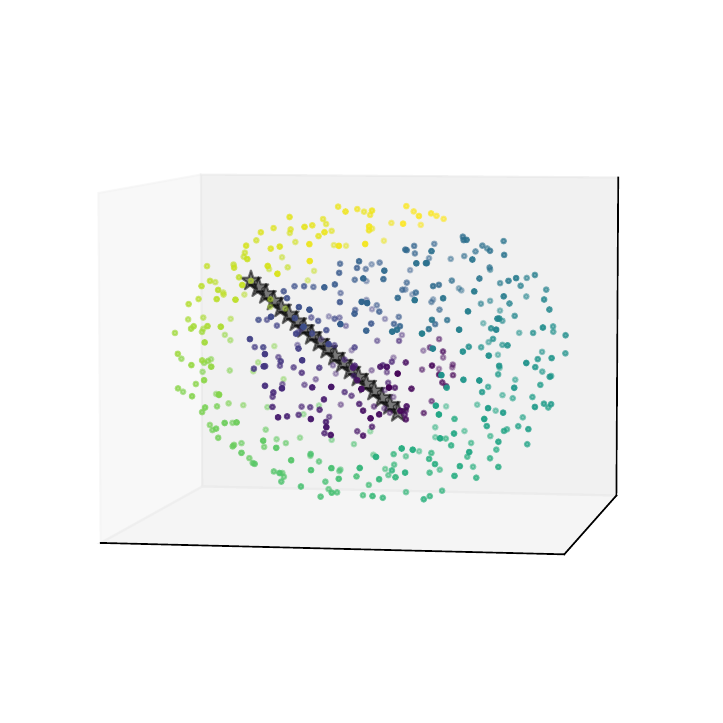} \\
        & 
   \end{tabular}
    \caption{\footnotesize{\textbf{Interpolation in Latent Space.} \emph{Colored points} are samples drawn from a Swiss roll (left) and their positions after forward optimization (right). The $\star$s are the generated sample $y^{(0)}$ in EFS (right), and their location after applying  Algorithm~\ref{alg:backward} (left). The straight line in the latent space (right) is transformed into a curved trajectory (left) that aligns with the data distribution. We observe that the algorithm avoids generating data from no density regions. }}
    \label{fig:swiss_interpolation}
\end{figure}

\begin{itemize}
    \item[(i)] \emph{Forward optimization:}In Figure~\ref{fig:forward_appendix}, we observe that the distribution of data points becomes uniform over the unit ball. Recall that Corollary~\ref{cor:largem} proves this result in the asymptotic regime as \( n \to \infty \). Figure~\ref{fig:forward_appendix} shows that this asymptotic result provides a good approximation even when \( n = 500 \).
    \item[(ii)] \emph{Backward optimization:}We assert that EFS can generate new samples from the Swiss roll dataset using a given set of i.i.d. samples. While Theorem~\ref{thm:backward} establishes that EFS draws samples from the data distribution in the mean-field regime, Figure~\ref{fig:backward_appendix} supports this result in the non-asymptotic setting.
  \item[(iii)] \emph{Interpolation:} Figure~\ref{fig:swiss_interpolation} illustrates how backward optimization can generate new samples from the Swiss roll dataset via interpolation. Similar to the Gaussian mixture case shown in Figure~\ref{fig:mixture_interpolation}, we observe that EFS respects the data density and avoids generating samples in low-density regions.
\end{itemize}

\begin{figure}[h]
  \centering
\begin{tabular}{|M{2.5cm}|| M{9cm}|}
\hline
Generated Images & \includegraphics[width=\linewidth]{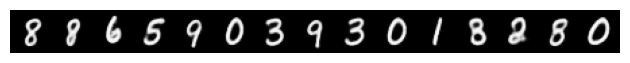} \\
\hline
Closest Image & \includegraphics[width=\linewidth]{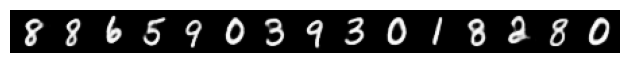} \\
\hline
    Animation (see with Adobe) & \animategraphics[autoplay,loop,width=\linewidth]{5}{imgs/frame}{1}{2} \\
    \hline
\end{tabular}
\caption{\footnotesize{\textbf{Generative MNIST.}  
\emph{First row:} Images generated by EFS. These are obtained by applying forward and backward optimization in the latent space of an autoencoder, followed by decoding into the image space.  
\emph{Second row:} The closest decoded training samples \( \{x_i^{(0)}\}_{i=1}^n \) for each generated image, determined by Euclidean distance.  
\emph{Third row:} An animation illustrating subtle differences between the generated images and their nearest neighbors (use Adobe Reader to see).  
\emph{Note:} The minimum Euclidean distance between each generated image and its nearest neighbor is 0.75.
}} \label{fig:minist_new}
\end{figure}

\paragraph{MNIST: Generating New Samples.} 
EFS can generate new  MNIST images~\citep{lecun1998gradient}.   Theorem~\ref{thm:backward} requires high exponents in the range \( d - 2 \leq m < d \). For MNIST with \( d = 784 \), this exponent leads to numerical overflow due to limited floating-point precision. To avoid this issue, we reduce the data dimension before applying EFS. Specifically, we use a convolutional autoencoder~\citep{masci2011stacked} to compress the data into a 15-dimensional latent space (see Appendix for architectural and training details). 
We then apply encoding $\to$ EFS $\to$ decoding to generate new samples. Figure~\ref{fig:minist_new} shows generated samples alongside their nearest neighbors (in Euclidean distance) from decoded original samples $x_1^{(0)}, \dots, x_n^{(0)}$. The generated digits exhibit stylistic variations distinct from their closest training examples, with minimum Euclidean distance of 0.7508. 
To emphasize subtle differences, we include an animated image that alternates between the generated sample and its nearest neighbor (open with Adobe Reader). See Table~\ref{table:params} for details.

\paragraph{MNIST: Interpolation Comparison.} Recall that in the previous experiment, we used an autoencoder for dimensionality reduction. Notably, autoencoders can generate samples with interpolation in the latent space. To highlight differences, we compare the quality of the autoencoder samples with those produced by EFS.

\begin{figure}[h]
  \centering
\begin{tabular}{|  M{13cm}|}
\hline
  \includegraphics[width=\linewidth]{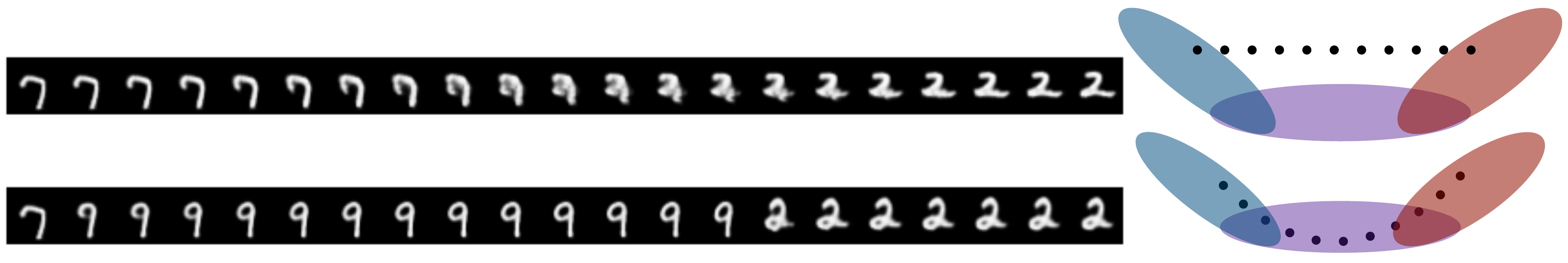}  \\
 \hline
 \end{tabular}  \caption{\footnotesize{\textbf{Interpolation Comparison. }  In the interpolation between images 7 and 2, the intermediate autoencoder outputs do not resemble valid digits (1st row), whereas our method produces a transition via $7 \rightarrow 9 \rightarrow 2$ (2nd row). On the right, we visualize the interpolation paths cross low density regions for autoencoder, while moves along the underlying digit clusters for EFS. Moreover, for EFS, it can be observed that different images are generated for each digit, illustrating that interpolation paths  evolve smoothly within clusters.}} \label{fig:minist_inter}
\end{figure}

For EFS, we generate new samples by linearly interpolating between uniformly distributed samples after forward optimization as 
$
y^{(k)} = (1 - t) x_i^{(k)} + t x_j^{(k)}
$. Then, we apply the backward optimization to \( y^{(k)} \). Similarly, autoencoder samples are obtain by decoding $(1-t) x_i^{(0)} + t x_j^{(0)}$ where $x_i^{(0)}$ are encoded samples in the latent space.  Figure~\ref{fig:minist_inter} shows generate samples with these two different strategies.
Observe interpolation in the autoencoder's latent space often leads to unrealistic outputs that do not resemble valid digits. As illustrated in the cartoon of Figure~\ref{fig:minist_inter}, such interpolation may cross regions of low data density. In stark contrast, EFS effectively avoids generating samples from these low-density regions, as similarly observed for Gaussian mixtures in Figure~\ref{fig:minist_new}.

 \begin{table}[h!]
\begin{center}
\begin{tabular}{l|ccccccc}
\toprule
 & $\gamma$ & $k$ & $T$ & $\beta$ & $\epsilon$ & $s$ & $n$\\
\midrule
Gaussian mixtures &  $0.1$ & $31$ & $300$ & $0.1$  & $0.001$ & $1$ & $400$\\
MNIST       & $0.05$ & $120$ & $300$ & $0.01$  & $0.001$ & $d-2$ &$15000$  \\
 Swiss roll       & $0.05$ & $120$ & $300$ & $0.1$  & $0.001$ & $d-2$ &$500$ \\
\bottomrule
\end{tabular}
\end{center} 
\caption{\footnotesize{Parameters of EFS }}\label{table:params}
\end{table}
\section{Related works}
\paragraph{Data generation with estimation.}
 Generative models traditionally rely on transporting simple to complex distributions. A classical example is the Box–Muller transform: given independent random variables $\theta \sim \text{Uniform}[0, 2\pi]$ and $r \sim \text{Exponential}$, the random vector $v = r(\cos \theta, \sin \theta)$ follows a standard two-dimensional Normal distribution~\citep{box1958note}. This illustrates a fundamental idea: sampling from a complex distribution can be achieved via a nonlinear transformation of samples drawn from a simpler distribution.

Modern generative models adopt this principle using function approximation: Given latent variable $x$ sampled from a Gaussian distribution, the goal is to find a parametric function $f_\theta$ such that the distribution of $f_\theta(x)$ approximates the data distribution. In generative adversarial networks (GANs), $f_\theta$ is implemented by a neural network and trained via adversarial objectives~\citep{goodfellow2014generative}. In normalizing flows, $f_\theta$ is a sequence of invertible and differentiable transformations optimized via maximum likelihood~\citep{rezende2015variational}. Score-based diffusion models, gradually transform  data using estimated score functions of diffusion density~\citep{song2020score}. In contrast to these methods, our proposed approach computes a non-linear transformation using a set of interacting points, thereby avoiding the estimation of transport map $f_\theta$.

\paragraph{Varational perspective towards generative models.}
Despite its distinct mechanism, EFS shares conceptual parallels with  diffusion models. Both can be interpreted as discretizations of different \emph{"Wasserstein gradient flows"}~\citep{ambrosio2008gradient}. This connection arises from a common variational principle: many simple distributions from which i.i.d. samples are easily drawn, such as the Gaussian distribution or the uniform distribution on a sphere, can be viewed as equilibrium states of physical systems with minimum energy. For example, Gaussian distribution is the minimizer of the negative entropy functional over the space of probability measures~\citep{jordan1998variational}, while the uniform distribution on the unit sphere minimizes a Riesz-type interaction energy~\citep{frank2025minimizers}. More generally, such distributions can be described as
\begin{align} \label{eq:min_p}
    p^* = \arg\min_{p \in \Omega} F(p), \quad \Omega := \{ \text{probability measures over } \mathbb{R}^d \},
\end{align}
where $F$ is an appropriate energy functional.

This variational formulation provides a principled mechanism to transport any distribution to the minimum energy state by gradient descent on $F$ in the space of probability densities. Diffusion models, optimize $F(p) = \int \log(p(x)) p(x) dx$ using Wasserstein gradient flow~\citep{jordan1998variational}. More interesting observation is that the reverse of gradient flow which can transport Gaussian distribution to any distribution. For entropy, inverse Wasserstein gradient flow is implemented by the Kolmogorov backward equation~\citep{kolmogoroff1931analytischen}, enabling transformation from a Gaussian to arbitrary target distributions. This key result forms the basis of modern generative diffusion models~\citep{sohl2015deep,ho2020denoising,song2020score}, where the reverse-time stochastic process includes a drift term proportional to the score function of intermediate densities~\citep{anderson1982reverse}. However, estimating this score function for arbitrary distributions remains statistically and computationally challenging. In this vein, \citet{wibisono2024optimal} recently proved that estimating the score function of sub-Gaussian distributions with Lipschitz-continuous scores suffers from a curse of dimensionality in the required sample complexity. \citet{songcryptographic} further demonstrated that, under lattice-based cryptographic hardness assumptions, score estimation remains computationally intractable even when the sample complexity is polynomial in the relevant parameters.To circumvent these challenges, our method replaces the entropy-based potential with a Riesz-type interaction energy and optimizes it directly over discrete empirical measures.

Inspired by Formulation~(\ref{eq:min_p}), one may attempt to optimize functionals of the probability measure \(p\) that quantify its distance to a target distribution \(p^*\). For example, the celebrated work of \citet{arjovsky2017wasserstein} introduce the Wasserstein GAN, which optimizes the Wasserstein-1 distance between the model distribution \(p\) and the target distribution \(p^*\) by solving its Kantorovich–Rubinstein dual via an adversarial training objective over both the generator (which parameterizes \(p\)) and the critic (which approximates the dual). More closely related to gradient flows on the space of probability measures, \citet{arbel2019maximum} consider the Wasserstein gradient flow of the Maximum Mean Discrepancy (MMD) functional and provide particle-based implementations. Additionally, \citet{mroueh2019sobolev} propose Sobolev Descent, deriving a particle algorithm from the Sobolev integral probability metric that transports samples along smooth descent paths. More recently, there is been a surge in using the seminal proximal point method of Jordan,
Kinderlehrer, and Otto (JKO)~\citep{jordan1998variational}, to discretisize the backward Wasserstein gradient flows over the chosen energy function $F$~\citep{mokrov2021large,alvarez2021optimizing,fan2022variational,bonet2022efficient,bunne2022proximal,altekruger2023neural}. Central to almost all of these works, is the estimation of the transport map in JKO proximal step with input convex neural networks~\citep{amos2017input}. In contrast to these works, our approach does not require parameterizing the transport map, the functionals that solve the variational formulations, or the underlying ODE/PDE induced by the continuity equation. Instead, it generates data via direct (inverse) gradient descent on a finite set of points without function estimation.

\paragraph{Particle gradient descent.} 
\cite{nitanda2017stochastic} introduce "particle gradient descent" for optimizing a given energy function $F$ over sparse measures as 
\begin{align}
    \min_{x_1, \dots, x_n } F \left( \frac{1}{n} \sum_{i=1}^n \delta(x_i) \right), \quad \delta(x): \text{the Dirac measure at $x$.}
\end{align}
Particle gradient descent is the gradient descent optimizing the location of $x_1, \dots, x_n$. 
Single layer neural networks can be viewed as particle gradient descent on a specific energy functions. Motivated by this connection,  \cite{chizat2018global} study the connection between gradient descent on particles and gradient flow in the space of probability measure equipped with Wasserstein-2 metric in asymptotic regime $n\to \infty$. \cite{daneshmand2023efficient} establishes a non-asymptotic convergence analysis for particle gradient descent on displacement convex functions. While primary focus of these studies is on analyzing single-layer neural networks, we leverage (inverse) particle gradient descent to develop an estimation-free generative method.  

\paragraph{Physics-inspired methods.}
 \citet{xu2022poisson} developed a generative model that maps samples from a uniform distribution over an infinite-radius hemisphere to an arbitrary target distribution. Their model relies on estimating the \emph{Poisson vector field} parameterized by neural networks. This idea, primarily inspired by physical systems—especially electrostatic theory—led to further developments in subsequent works~\citep{xu2023pfgm++,kolesov2025field}. 

To cope with the challenges of sampling from an infinite-radius hemisphere, \citet{xu2022poisson} simulate the corresponding ODE by perturbing the training data and then estimate the negative normalized field from these perturbed samples. This contrasts with our approach, in which the primitive distribution is uniform on a finite-dimensional compact manifold (the sphere), and is therefore easy to sample from. Although we also transport samples from a uniform distribution to an arbitrary data distribution, our transport mechanism does not rely on function estimation; rather, it is purely based on (inverse) gradient descent over a finite set of points.

\section{Future works}
We demonstrate, both theoretically and experimentally, that it is possible to transport a uniform distribution to a target distribution without estimating a score function—using only i.i.d. samples from the target. Our method, Estimation-Free Sampling (EFS), avoids both noise injection and function estimation. Instead, it introduces interactions between samples by optimizing an energy functional, thereby shaping the empirical distribution of samples. EFS lies at the intersection of three foundational fields—mathematical optimization, potential theory, and generative AI —offering rich opportunities for interdisciplinary research.
\paragraph{Optimization for Sampling.} While most generative models are built upon stochastic differential equations, EFS is purely a \emph{deterministic optimization method}, opening new avenues for the optimization community to use powerful optimization techniques—such as accelerated methods, higher-order optimization methods, and efficient stochastic techniques—to generate data.

Notably, EFS involves both convex and non-convex optimization. Linking the non-convex gradient descent dynamics to a well-studied Wasserstein gradient flow, we establish asymptotic convergence guarantees. However, understanding the behavior of gradient descent in non-asymptotic regimes remains an open challenge. We believe the introduced asymptotic results provides a solid foundation for future non-asymptotic analyses.

\paragraph{Potential Theory.}
EFS induces attractive-repulsive interactions between training samples using a well-known potential function studied extensively in fractional potential theory~\citep{shu2021generalized, frank2025minimizers2, duerinckx2020mean}. By linking generative modeling to this rich theoretical framework, we gain access to powerful analytical tools for understanding and designing generative models. In particular, our analysis leverages results on Wasserstein gradient flows of attractive-repulsive energies~\citep{duerinckx2020mean}, as well as variational analysis of these energy functionals~\citep{carrillo2023radial, frank2025minimizers}. This bridge between theory and practice opens new directions for developing principled and reliable generative models with theoretical guarantees.

Potential theory can design new interaction potentials tailored to data generation and practical applications. As noted in our experiments, one major challenge was the choice of the power 
$s$ in the potential function, which led to numerical instability in high dimensions. Investigating alternative potentials that avoid such issues—particularly those that do not involve dimension-dependent blow-up—may help resolve computational challenges of EFS.

\paragraph{Generative AI.}
EFS opens up several promising directions for future research, particularly in scaling to large-scale machine learning applications. A key practical challenge lies in the quadratic time complexity of the forward optimization step with respect to the number of training samples. We conjecture that stochastic optimization techniques
  could mitigate this issue and significantly reduce computational cost.

\bibliographystyle{unsrtnat}
\bibliography{ref} 
 \appendix
\begin{center}
    \Large{\textbf{Appendix}}
\end{center}
\section{Algorithm} \label{section:algorithm}

For completeness, we present the EFS algorithm in \Cref{alg:sample} and its forward and backward optimization subroutines in \Cref{alg:forward} and \Cref{alg:backward_app}, respectively. Remarkably, even when the distribution is uniform over a high-dimensional ball, the samples are effectively drawn uniformly from the unit sphere, which is denoted by $\mathbb{S}^{d-1}$, as almost all the volume concentrates near the boundary \citep{ball1997elementary}.   
\begin{algorithm}[H]
\caption{Estimation Free Sampling (\textit{EFS})} \label{alg:sample}
\KwIn{I.I.D samples $\{x_1^{(0)},\dots, x_n^{(0)}\}$}

\KwSty{}{\textbf{Parameters}: Step size $\gamma$, number of forward iterations $k$, learning rate of backward (proximal step) $\beta$, number of iterations for each backward proximal step $T$, potential parameters  $s$ and $\epsilon$}

\smallskip  

\tcc{Forward the training data to the sphere by gradient descent}

\KwSty{}{\textbf{Set} $\{x_0^{(j)},\dots, x_n^{(j)}\}_{j=1}^k \leftarrow $ \texttt{Forward}$(\{x_1^{(0)},\dots, x_n^{(0)}\}, \gamma, k, s, \epsilon)$}

\smallskip  

\tcc{Draw a new random point uniformly from the sphere}
\KwSty{}{\textbf{Set} $c = \frac{1}{n} \sum_{i = 1}^n x_i^{(k)}, r = \frac{1}{n} \sum_{i = 1}^n \| c - x_i^{(k)}\| $}

\KwSty{}{\textbf{Draw} $\nu \sim \mathbb{S}^{d - 1}(c, r)$} \hfill \tcp{Sphere with center $c$ and radius $r$}

\smallskip  

\tcc{Backward the new data point to the original space}
\KwSty{}{\textbf{Set} $y^{0} \leftarrow $ \texttt{Backward}$(\{x_1^{(j)},\dots, x_n^{(j)}\}_{j=1}^k, y^{(k)}, \gamma, k, \beta, T, s, \epsilon)$}

\smallskip 

\KwOut{Generated sample $y^{(0)}$}
\end{algorithm}

\begin{algorithm}[thp]
\caption{Forward Optimization (\textit{Forward})} \label{alg:forward}
\KwIn{Training data $\{x_1^{(0)},\dots, x_n^{(0)}\}$}

\KwSty{}{\textbf{Parameters}: step size $\gamma$, number of forward iterations $k$, potential parameters $s$ and $\epsilon$}

\For{$j \leftarrow 0$ \KwTo $k-1$}{
    \For{$i \leftarrow 1$ \KwTo $n$}{
    $\Delta_i = \tfrac{1}{n - 1} \sum_{a \in [n], a \neq i} \nabla W_\epsilon^{(s)}(x_i^{(j)} - x^{(j)}_a)$
    
    $x_i^{(j + 1)} = x_i^{(j)} - \gamma  \Delta_i $
    }
}
\KwOut{$\{x_0^{(j)},\dots, x_n^{(j)}\}_{j=1}^k$}
\end{algorithm}

\begin{algorithm}[H]
\caption{Backward Optimization (\textit{Backward})} \label{alg:backward_app}
\KwIn{Data $\{x_1^{(j)},\dots, x_n^{(j)}\}_{j=1}^k$, new sample $y^{(k)}$}

\KwSty{}{\textbf{Parameters}: Step size $\gamma$, number of forward iterations $k$, learning rate of backward (proximal step) $\beta$, number of iterations for each backward proximal step $T$, potential parameters  $s$ and $\epsilon$}

\KwSty{}{\textbf{Set}: $j=k$ }

\While{$j\geq 0$}{
    \textbf{Set}: $v_0 = y^{(j)}$

    \For{$t \leftarrow 0$ \KwTo $T$}{
    $\nabla = \tfrac{\gamma}{n} \sum_{i \in [n]} \nabla W_\epsilon^{(s)}(v_t - x^{(j)}_i)$
    
    $\Delta = v_t - y^{(j)} - \nabla $
    
    $v_{t+1} = v_t - \beta  \Delta $
    }
    $j = j-1$ and $y^{(j)} = v_T$
    } 
\KwOut{$y^{(0)}$}
\end{algorithm}
\section{Preliminaries} \label{sec:background}
\paragraph{Notations.} $\mathbb{S}^{d-1}(c,r)$ denotes sphere with center $c \in \R^d$ and radius $r$. $\Omega$ denotes the set of probability measure over $\R^{d}$ with Wasserstein-2 metric, which is denoted by $\mathcal{W}_2(\mu,\nu)$ where $\mu,\nu\in \Omega$. Given vector function $v: \R^d\to \R^d$, its \emph{divergence} is defined as $\mathrm{div}(v) = \sum_{i=1}^d \frac{d v}{d x_i} $. Suppose $E: \Omega \to \R$. Then, the \emph{first variation} of $E$ with respect to $\mu$ is denoted by $\tfrac{dE}{d\mu}$~\citep{santambrogio2017euclidean}. Function $f: \R^d \to \R^d$ transport $\mu \in \Omega$ to $\nu\in \Omega$ if for a random vector $x \in \R^d$ drawn from $\mu$, $f(x)$ has density $\nu$. Recall $W^{(s)}_{\epsilon}$ is the potential function defined as 
\begin{align}
W^{(s)}_{\epsilon}(x - y) = \frac{\|x - y\|^2}{2} +  \frac{1}{s(\|x - y\|^2+\epsilon)^{s/2}}.
\end{align} For simplicity, we use the shorthand notation $W^{(s)} = W^{(s)}_0$ and \( W^{(s)} \) by \( W \). 
The sign $*$ denotes the standard convolution as  \begin{align} \label{eq:conv}
        f * g (y) = \int f (y-x) g(x) dx 
    \end{align}
 $\| f \|_{L^2}$ is the $L_2$ functional norm defined as 
 \begin{align}
     \| f\|_{L_2}^2 = \int \| f (x)\|^2 dx \label{eq:L2}
 \end{align}

\paragraph{Weak/Strong convergence.} 
A sequence of measures $\mu^{(n)} \in \Omega$ is said to converge weakly to $\mu$ if, for all bounded continuous test functions $f: \R^d \to \R$, the integrals $\int f(x) \mu^{(n)}(x) \, dx$ converge to $\int f(x) \mu(x) \, dx$. 
A stronger notion is convergence in the $L^2(\nu)$ norm, which requires that
\[
\lim_{n \to \infty} \int \| \mu^{(n)}(x) - \mu(x) \|^2 \nu(x) dx = 0.
\]
 
\paragraph{MMD.} 
Define $\mmd{}:\Omega \times \Omega \to \R_+$ as
\begin{align}
\mmd^2(\mu,\nu) = \int K(x-y) (\mu(x)-\nu(x))dx (\mu(y)-\nu(y))dy, \quad \textrm{with } K(\Delta) \coloneq \frac{1}{s\|\Delta\|^{s}}\numberthis{eq:mmd_def}
\end{align}

where $\mmd$ denotes the Maximum Mean Discrepancy between two measures $\mu$ and $\nu$.

Suppose that $\widehat{K}$ and $\widehat{\Delta}$ denote the Fourier transforms of the functions $K(x)$ and $\Delta(x) \coloneqq \mu(x)-\nu(x)$, respectively. Viewing \mmd{} as a convolution and Plancherel's theorem allow us to write \mmd{} as~\citep{shu2021sharp}
\begin{align}
    \mmd^2(\mu,\nu) = \int \widehat{K}(w) |\widehat{\Delta}(w)|^2 dw, \numberthis{eq:definite}
\end{align}
where $\widehat{K}(w) = C \| w\|^{-d+s}$ with a constant $C$ depending only on $d$ and $s$~\citep{frank2025minimizers}. Substituting the Fourier transform into the equation above establishes that \( \mmd(\mu, \nu) = 0 \) implies \( \mu = \nu \). Notably, it is sufficient for the Fourier transform \( \widehat{K} \) to be strictly positive almost everywhere to guarantee that \( K \) is a \emph{universal kernel}~\citep{gretton2012kernel}. 

To avoid potential confusion, note that the $\mmd{}$ associated with the kernel $K$ defined above is not a metric. While machine learning often relies on positive semi-definite kernels to ensure that $\mmd{}$ defines a valid metric, the kernel $K$ is not positive semi-definite. As a result, the corresponding $\mmd{}$ does not obey the triangle inequality. 

\paragraph{Gradient dominance of $E$.} We establish an important property of the energy function $E$. Define functional $F: \Omega \to L_2$ as \begin{equation} 
F(\mu)(y) = \nabla \frac{d E}{d\mu} = \int W(y-x) \mu(x) dx.
\end{equation} Remarkably, $\mu$ is a steady state of the Wasserstein gradient flow on $E$ if $F(\mu) = 0$. The next theorem represent $\mmd{}$ using $F$. 
\begin{theorem} \label{thm:derivative_convergence}
        Suppose that $\mu \in \Omega$ and $\nu \in \Omega$ have the same first-moment, then 
        \[
        \| F(\mu) - F(\nu) \|_{L_2}^2 =\mmd{}^2(\mu,\nu)
        \]
        holds for $s=d-2$.
\end{theorem}
 An application of the above theorem recovers the result of \citet{carrillo2023radial}: all steady states $\mu$ satisfying $F(\mu) = 0$ are global minimizers of $E$ (up to translation). More precisely, the theorem implies that for all $\mu, \nu$ such that $F(\mu) = F(\nu) = 0$, the following holds:
\begin{align}
   0 = \| F(\mu) - F(\nu) \|_{L_2}^2 = \mmd{}(\mu,\nu).
\end{align}
Beyond recovering existing results, the theorem will also be used to analyze the backward optimization for EFS in Section~\ref{sec:backward_app}.
\begin{proof}
    Let $\widehat{f}(w)$ denote the Fourier transform of function $f(x)$. Define $\widehat{\Delta} \coloneqq \widehat{\mu} - \widehat{\nu} $ which is equivalent to the Fourier transform of $\Delta \coloneqq \mu-\nu$. As discussed in Section~\ref{sec:background}, $\mmd$ can be written as
    \begin{align}  \mmd^2(\nu,\mu) =  \int \widehat{K}(w) |\widehat{\Delta}(w)|^2 dw \label{eq:mmd_fourier}
    \end{align}
    According to the definition, 
    \begin{align}
        \nabla W(z) = \nabla K (z) + \mathrm{iden}(z), \quad \mathrm{iden}(z):= z 
    \end{align}
    Replacing the above formula into the definition of $F$ obtains
    \begin{align}
         \|  F(\mu) - F(\nu)\|_{L_2}^2  = \| (\nabla K + \mathrm{iden}) * \mu- (\nabla K + \mathrm{iden}) * \nu \|_{L^2}^2,  
    \end{align}
    where $*$ denotes convolution defined in \eqref{eq:conv} and $L_2$ is the norm 2 for functions defined in \eqref{eq:L2}.  
    It is easy to check that 
    \begin{align}
        (\mathrm{iden} * \mu)(y) & = y \underbrace{\int \mu(x) dx}_{=1} - \int x \mu (x) dx \\
        & = y \underbrace{\int \nu(x) dx}_{=1} - \underbrace{\int x \mu (x) dx}_{\mathbb{E}_{\mu}[x]}  \\
        & = y \int \nu(x) dx - \underbrace{ \int x \nu (x) dx}_{\mathbb{E}_{\nu}[x]} = (\mathrm{iden}*\nu)(y). 
    \end{align}
 The last equation holds because the first moments of $\mu$ and $\nu$ are equal. This observation allows us to significantly simplify the expression for the quantity of interest as follows:
  \begin{align}
      \| F(\mu) - F(\nu)\|_{L_2}^2 = \| \nabla K * \mu - \nabla K*\nu\|^2_{L_2}
  \end{align}
Recall two fundamental properties of Fourier transform as: $\begin{cases}
    \widehat{f*g} = \widehat{f}\widehat{g} \\
    \widehat{\partial f} = iw \widehat{f}
    \end{cases}$
. These properties together with 
    Parseval's theorem
     yield  
    \[
    \| F(\mu) - F(\nu)\|_{L_2}^2 
     = \int \| w \|^2\left(\widehat{K} (w)\right)^2 |\widehat{\Delta}(w)|^2 dw \numberthis{eq:fourier_parseval}
    \]
    For $s=d-2$, it is easy to check that \begin{equation}\|w\|^2 (\widehat{K}(w))^2 =  \widehat{K}(w) \label{eq:s_choice_consequence}\end{equation} holds given $\widehat{K}(w) =C\|w\|^{-2}$ \citep{frank2025minimizers}. Combining all the results completes the proof:
    \begin{align}
        \| F(\mu) - F(\nu)\|_{L_2}^2  & \stackrel{\eqref{eq:fourier_parseval}}{=}  \int \| w \|^2\left(\widehat{W} (w)\right)^2 |\widehat{\Delta}(w)|^2 dw \\
        & \stackrel{\eqref{eq:s_choice_consequence}}{=} \int \widehat{W}(w) |\widehat{\Delta}(w)|^2 dw \\
    & \stackrel{\eqref{eq:mmd_fourier}}{=} \mmd{}^2(\mu,\nu)
    \end{align}
\end{proof}
\paragraph{A convergence result for ODEs.}
Consider the following two differential equations 
\begin{align}
    \begin{cases}
        \frac{dy}{dt} = F_t(y_t) \\
        \frac{dy^{(n)}}{dt} = F_t^{(n)}(y^{(n)})
\end{cases}, \quad y_0 = y_0^{(n)}
\end{align}
where $F_t, F_t^{(n)}: \R^d \to \R^d$ and $y_t,y_t^{(n)}\in \R^d$. If $F_t^{(n)}$ converges to $F_t$ as $n\to \infty$, then $y_t^{(n)}$ converges to $y_t$ in interval $t \in [0,T)$. 
\begin{lemma} \label{lemma:ode_convergence}
    Define 
    \[
    \epsilon_n \coloneq \int \| F_t(y) - F_t^{(n)}(y)\|^2 d y.\]
    Suppose $F_t(y)$ is almost surely $L$-Lipschitz in $y$. If $\lim_{n\to \infty}\epsilon_n =0$, then $\int_0^T \| y_t - y_t^{(n)}\|^2dt$ converges to $0$ as $n\to \infty$. 
\end{lemma}
\begin{proof}
    We start by writing down ODEs in the integral form and consider their difference,
    \[
    \|y^{(n)}(t) - y(t)\| & = \left\|\int_{0}^t \mleft[ F_s^{(n)}(y^{(n)}) - F_s( y) \mright] d s \right\| \\
    & = \left\|\int_{0}^t \mleft[ F_s^{(n)}(y^{(n)}) - F_s( y^{(n)}) \mright] d s + \int_{0}^t \mleft[ F_s(y^{(n)}) - F_s( y) \mright] d s \right\| \\
    & \leq t \; \sup_{y} \mleft\| F_s^{(n)} (y) - F_s (y) \mright\| + L \int_{0}^t \mleft\|y^{(n)}(s) - y(s) \mright\| ds \\
    & \leq t \epsilon_n +  L \int_{0}^t \mleft\|y^{(n)}(s) - y(s) \mright\| ds \numberthis{eq:take_sup_from_this}
    \]
    The last inequality was followed by convergence of $F^{(n)}_s$ to $F_s$, and boundedness of $L^\infty$ norm by $L^2$, the first term is converging to zero, and bounded by $\epsilon_n  \xrightarrow[]{n \rightarrow \infty} 0$.

    Define $\Delta_n(t) \coloneq \sup_{s \leq t} \mleft\| y^{(n)}(s) - y(s) \mright\|$. By taking supremum from \eqref{eq:take_sup_from_this},
    \[
    \Delta_n(t) \leq t \epsilon_n +  L \int_{0}^t \Delta_n(s) ds.
    \]
    In turn, applying Grönwall's inequality yields,
    \[
    \Delta_n(t) \leq t \epsilon_n e^{L t} \xrightarrow[]{n \rightarrow \infty}
     0.
    \]
    Thus,
    \[
    \mleft\| y^{(n)}(s) - y(s) \mright\| \xrightarrow[]{n \rightarrow \infty}
     0, 
    \]
    uniformly for all $s \leq t$ and proof is completed.
\end{proof}

\paragraph{Continuity equation and transporting distributions.}
Let $v_t: \R^d\to \R^d$ be a general vector field. $\mu_t \in \Omega$ is a solution of continuity equation associated with $v_t$ if it obeys 
\begin{align}
    \frac{d \mu_t}{dt} = - \mathrm{div}(\mu_t v_t)
\end{align}
Given the vector field, we define the following ODE 
\begin{align}
\frac{dy_t}{dt} = v_t(y_t)
\end{align}
The above ODE transports $\mu_0$ to $\mu_t$ as stated in the following lemma. 
\begin{lemma}[Lemma 8.1.6. of \citep{ambrosio2008gradient}] \label{lemma:transport}
Suppose that the vector field $v_t$ obeys the following 3 conditions 
\begin{itemize}
    \item[(1)] $\int |v_t(x)|d\mu_t(x)dx< \infty$. 
    \item[(2)] For every compact subset $B\subset \R^d$, $\sup_{x \in B} |v_t(x)| < \infty$. 
    \item[(3)] $v_t(x)$ is Lipschitz  
\end{itemize}
If $y_0$ is a random variable with distribution $\mu_0$, then $y_t$ is a random variable with distribution $\mu_t$ for a finite $t$.
\end{lemma}

\newpage
\section{Proof of Theorem~\ref{thm:backward}} \label{sec:backward_app}
 
\begin{figure*}[h!]
    \centering
\begin{tabular}{M{4.5cm} M{5.5cm}  M{4cm}}
        \hline 
        \hline
        \includegraphics[width=0.3\textwidth]{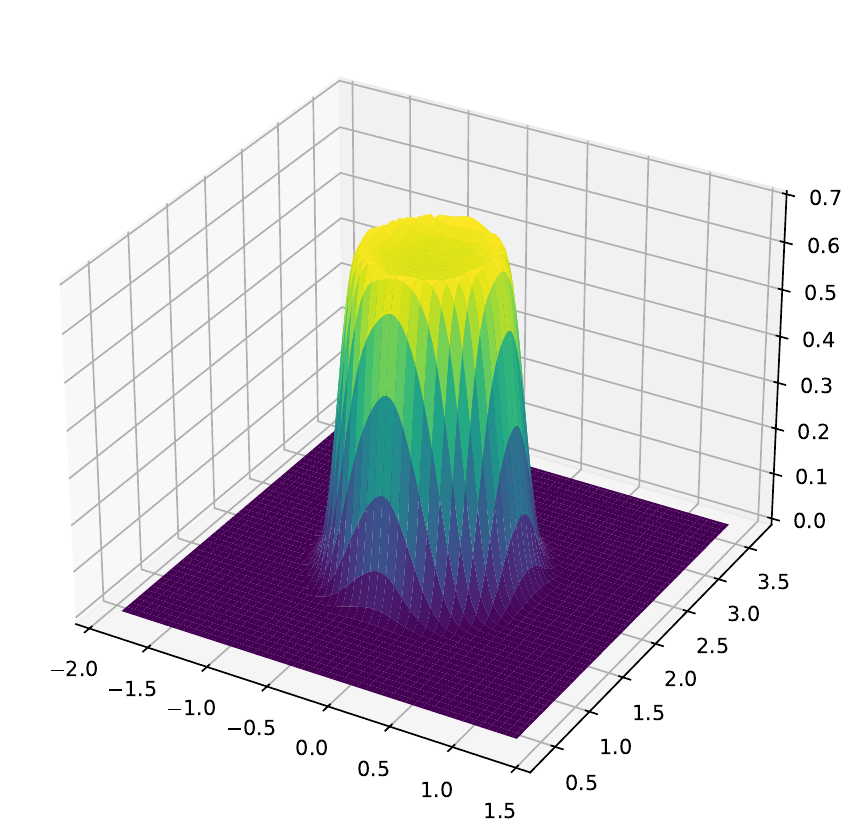} & \begin{multline*}
    \textrm{(almost) uniform}  \sim y_t \stackrel{T}{\Longrightarrow} y_0 \sim \textrm{data} \\
T(y_t) = \int_{t}^0 \int \nabla W(y-x)\mu_t(dx)dt        \end{multline*} 
        & \includegraphics[width=0.3\textwidth]{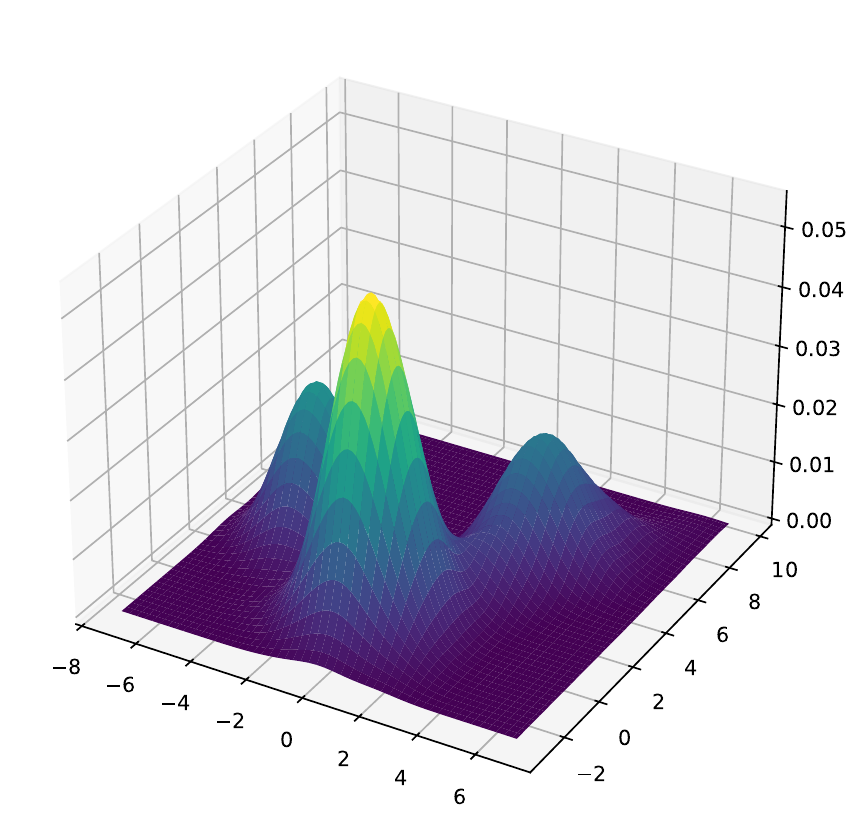}    \\
\includegraphics[width=0.3\textwidth] {imgs/cir_new.pdf} & 
\begin{multline*}\textrm{uniform}  \sim y_t \stackrel{T^{(n)}}{\Longrightarrow} y_0^{(n)} \sim \textrm{(almost) data} \\
T^{(n)}(y_t) = \int_{t}^0 \int \nabla W(y-x) \mu_t^{(n)}(dx) dt
        \end{multline*}
&\includegraphics[width=0.3\textwidth] {imgs/mixture_new.pdf} \\
\hline
\hline
    \end{tabular}
    \caption*{\textbf{Proof sketch for \Cref{thm:backward}.} Given the solution to the continuity equation~\eqref{eq:continuity}, we define the map \( T \) above, which provably transports $\mu_t$ to data distribution, as stated in Lemma~\ref{lemma:backward}. Recall that Theorem~\ref{thm:frank} states that \( \mu_t \) converges to the uniform distribution, from which new samples can be drawn.  $T$ is not implementable as it requires \( \mu_t \), the solution of the continuity equation~\eqref{eq:continuity}. We show that the empirical distribution \( \mu_t^{(n)} \), defined over the point set \( \{x_1(t), \dots, x_n(t)\} \), yields a transport map \( T^{(n)} \) (as defined above) that converges to \( T \) as \( n \to \infty \). The $\star$ symbol in the second row indicates \( y_t \) (left) and \( y_0^{(n)} \) (right). Colored points represent \( \{x_1(t), \dots, x_n(t)\} \) on the left and \( \{x_1(0), \dots, x_n(0)\} \) on the right.
  }
\end{figure*}
\paragraph{Recap.}
Recall that $\mu_t^{(n)}$ defined as $\mu_t^{(n)} = \frac{1}{n} \sum_{i=1}^n \delta_{x_i(t)}$ where $\delta$ is the Dirac measure and $x_i(t)$ obeys 
\begin{align} 
\frac{dx_i}{dt} = - \frac{1}{n} \sum_{j \neq i} \nabla W^{(s)}(x_i - x_j).
\end{align}
Using $x_i(t)$, we define $y_t^{(n)}$ which obeys
\begin{align}
    \frac{dy^{(n)}}{d\bar{t}} = F_t^{(n)}(y_t^{(n)}), \quad F_t^{(n)}(y)\coloneq\int \nabla W\left(y - x\right)\mu_t^{(n)}(x)dx, \numberthis{eq:ode_emp}
\end{align}
where $d\bar{t}$ indicates that the process is reversed time~\citep{anderson1982reverse}.

    \paragraph{Reversed-time transport.}     
    Recall $\mu_t$ as the solution of continuity equation for Wasserstein-2 gradient flow:
    \begin{align}
    \frac{d\mu}{dt} = \mathrm{div}\left( \mu(x) \int \nabla W^{(s)}(x - y) \mu_t(y) \, dy \right), \label{eq:gradient_flow_appen} 
    \end{align}
    Notably, $F_t(y) \coloneq \int \nabla W(y - x) \mu_t(x) dx$ represents the vector field associated with the above dynamics. Given this vector field, define the following ODE
    \begin{align}
        \frac{dy}{dt} = -  F_t(y_t), \quad \textup{where } F_t(y) \coloneq  \int \nabla W(y - x) \mu_t(x) dx \label{eq:ode_backward_gradientflow}
    \end{align}
    The above ODE transports $\mu_t$ backward to $\mu_0$ as stated in the following lemma.
    \begin{lemma} \label{lemma:backward}
        Let $y_0$ be a random vector drawn from $\mu_0$. Then, the distribution of $y_t$ is $\mu_0$. 
    \end{lemma}
The above result together with a simple change of variables obtains the inverse transport from $\mu_t$ to $\mu_0$. 
\begin{corollary}
    Consider the following reversed-time dynamics: 
    \begin{align}
        \frac{dy_t}{d\bar{t}} = F_t(y_t)
    \end{align}
    The above dynamics transports $\mu_t$ to $\mu_0$ as long as $\mu_t$ is not a steady state of \eqref{eq:gradient_flow_appen}.
\end{corollary}
\begin{proof}
    According to definition 
    \begin{align}
        \lim_{\epsilon\to 0} \frac{y_{t+\epsilon}-y_t}{\epsilon} = -F_t(y_t) \implies \frac{dy}{d\bar{t}} =\lim_{\epsilon\to 0} \frac{y_{t-\epsilon}-y_{t}}{\epsilon} = \lim_{\epsilon \to 0} F_{t-\epsilon}(y_{t-\epsilon})
    \end{align}
    where $F_{t-\epsilon}(y) = \int \nabla W(y-x)\mu_{t-\epsilon}(x) dx$. Lemma~\ref{lemma:Lipschitzness} implies $F_t$ is Lipschitz. Thus, invoking Lemma~\ref{lemma:ode_convergence} concludes the statement.
\end{proof}
 While the flow \( y_t \) transports \( \mu_t \) to the data distribution, its construction relies on access to \( \mu_t \), which is defined as the solution to a PDE. Since \( \mu_t \) is not feasible to compute, we cannot implement the reverse-time ODE~\eqref{eq:ode_backward_gradientflow} to recover \( y_0 \). However, Theorem~\ref{thm:backward}
 restated bellow ensures that EFS can reconstruct the backward dynamic as $n\to \infty$. 
\begin{repeatthm}{thm:backward}
    Assume $\mu_0^{(n)}$ converges to $\mu_0$ in Wasserstein-2 distance. 
    Suppose that \( y_t^{(n)} \) is a random variable with law \( \mu_t \), where \( \mu_t \) is the solution to the continuity equation~\eqref{eq:continuity}.
    Then, the distribution of \( y_0^{(n)} \)—obtained from the reversed-time ODE~\eqref{eq:backwardode},
    \[
    \frac{dy^{(n)}}{d\bar{t}} = \frac{1}{n} \sum_{i=1}^n \nabla W^{(s)}\left(y^{(n)} - x_i\right)
    \]
    converges to \( \mu_0 \) as \( n \to \infty \), for regular measures \( \mu_0 \) and \( \mu_t \), and for \( s = d - 2>0 \).
\end{repeatthm}
\begin{proof}
We prove that $y_0^{(n)}$ converges to $y_0$, whose distribution is $\mu_0$, according to Lemma~\ref{lemma:backward}. To establish this convergence, we use Theorem 1 of \citep{duerinckx2020mean}, which shows that $\mu_t^{(n)}$ converges to $\mu_t$ in \mmd{} as $n \to \infty$, namely
\begin{align}
    \lim_{n \to \infty} \mmd{}(\mu_t, \mu_t^{(n)}) = 0, \quad \text{for } \beta > 0.
\end{align}
As proven in the next lemma, $F_t(y)$ is Lipschitz. This Lipschitz continuity, together with the result above, ensures that the conditions in Lemma~\ref{lemma:ode_convergence} are satisfied. Thus, $y_0^{(n)}$ converges to $y_0$, as guaranteed by the lemma.
\end{proof}
\begin{lemma} \label{lemma:Lipschitzness}
    Recall vector function
    $F_t(y) = \int \nabla W^{(d-2)}(y-x) \mu_t(x) dx$ where $\mu_t$ is the solution of PDE~\eqref{eq:gradient_flow_appen} starting from $\mu_0$. Then, $F_t(y)$ is $L$-Lipschitz for $L=c_1 + c_2 (E(\mu_0)+1)^{\tfrac{2d+2}{d}}$ where constants $c_1$ and $s_2$ depend only on $d$. 
\end{lemma}
\begin{proof}
The first step is to cast Lipschitzness to the boundedness of the following integral 
\begin{align}
   \int \frac{1}{\|y-x\|^{d}} \mu_t(x)dx \leq L \implies F_t \text{ is $L$-Lipschitz }
\end{align}
Then, we use properties of Wasserstein gradient flow to prove 
\begin{align}
    \int \frac{1}{\|y-x\|^{d}} \mu_t(x)dx \leq c_1 + c_2 E(\mu_0), 
\end{align}
where $c_1$ and $c_2$ are constants that only depend on $d$. For the similarity, we introduce notation $a \lesssim b$ if there are constants $c_1$ and $c_2$ that only depends on $d$ such that $a \leq c_1 + c_2 b$ 

\emph{Step I. A Sufficient Condition for Lipschitz-ness.} According to the definition, we have

\begin{align}
   \|  F_t(y) - F_t(y')\|_2 & = \left\| \int \nabla K(y-x)\mu_t(dx) - \int \nabla K(y'-x) \mu_t(dx)  \right\|_2 \\
   & \leq \| (y-y') \int_{0}^1\nabla^2 K(\gamma y + (1-\gamma)y'-x) \mu_t(dx) d\gamma \|_2 \\
   & \leq \| y - y'\|_2 \max_{y} \|\int  \nabla^2 K(y-x) \mu_t(x)dx\|_2
   \end{align}
   To prove that the right side is bounded, we bound the norm of $\nabla^2 K$ as 
   \begin{align}
       \| \int \nabla^2 K(y-x) \mu_t(x)dx \| & = \left\| \int -\frac{I_d}{\| y-x\|^{d}} + (d)\frac{(y-x)(y-x)^\top}{\|y-x\|^{d+2}}\mu_t(dx) \right\| \\
       & \leq (d) \int \frac{1}{\|y-x\|^{d}} \mu_t(x)dx
   \end{align}
Thus, the above integral bounds the Lipschitz constant of $F_t$. 

\emph{Step II. Gradient flow property.}
Since $\mu_t$ is a gradient flow, the following holds 
\begin{align}
  E(\mu_t) \leq E(\mu_0)  \label{eq:flow_property}
\end{align}
We establish an important consequence of the above inequality, namely the following bound
\begin{align}
    \sup_{y} \left|\int \frac{1}{\|y-x\|^{d-1}} \mu_t(x) dx \right| \lesssim (1+E(\mu_0))^2  \label{eq:Lcondition}
\end{align} 
holds. Using Jensen's inequality, we have 
\begin{align}
    \int \frac{1}{\|x-y\|} d\mu_t(x)\leq \left( \int \frac{1}{\|x-y\|^{d-1}} d\mu_t(x)\right)^{\tfrac{1}{d-1}} 
\end{align}
Combining the last two bounds obtains 
\begin{align}
    \int \frac{1}{\|y-x\|^{d}}\mu_t(x)dx \leq \left|\int \frac{1}{\|y-x\|^{d-1}} \mu_t(x) dx \right| \left|\int \frac{1}{\|y-x\|} \mu_t(x) dx \right| \lesssim (1+E(\mu_0))^{\tfrac{2d+2}{d}}
\end{align}
The above equation concludes that $F_t$ is Lipschitz according to step I. To complete the proof, we need to prove equation~\eqref{eq:Lcondition}. 

We first establish a consequence of $E(\mu_t) \leq E(\mu_0)$ for $\mu_t$. 
    Let $\widehat{\mu}(w)$ denote Fourier transform of $\mu_t(x)$. It is easy to check that 
    \begin{align} \label{eq:norm1bound}
        |\widehat{\mu}(w)| \leq \int |\mu_t(x)|dx =1
    \end{align}
    Without of loss of generality, we can assume that $\int x \mu_t(x)dx = 0$. An application of Jensen's inequality yields
    \begin{align}
        \int \| x\|^2 d\mu_t(x) & =  \int \| x- \int y \mu_t(y)dy \|^2 d\mu_t(x) \\
        & \leq \int \| x - y\|^2 d\mu_t(x) d\mu_t(y) dx dy \\
        & \lesssim E(\mu_t) \lesssim E(\mu_0) \label{eq:norm_bound}
    \end{align}
According properties of Fourier transform, we have 
\begin{align}
    |\frac{\partial \widehat{\mu}}{\partial w_i \partial w_j}| & =  \int| x_i x_j| \mu_t(x) dx \\
    & \leq \frac{1}{2} \left( \int x_i^2 \mu_t(x) dx + \int x_j^2 \mu_t(x) dx  \right) \\ 
    & \leq \int \|x\|^2 \mu_t(x) \\
    & \stackrel{\eqref{eq:norm_bound}} {\leq}  E(\mu_0) \label{eq:norm2hessian}
\end{align}
Gagliardo–Nirenberg interpolation inequality yields 
\begin{align}
    \| \widehat{\mu}\|_{L^1} & \lesssim (\| \partial^2 \widehat{\mu} \|_{L^\infty}+1) (\| \widehat{\mu} \|_{L^\infty}+1) \\
    & \stackrel{\eqref{eq:norm2hessian}}{\lesssim} (1+E(\mu_0)) (\| \widehat{\mu} \|_{L^\infty}+1) \\
    & \stackrel{\eqref{eq:norm1bound}}{\lesssim} (1+E(\mu_0))^2 \label{eq:L1boundfourier}
\end{align}
We will use the above bound to complete the proof.

Recall the Fourier transform of the radial function $\| x\|^{-s}$  is 
$C \|w\|^{-d+s})$ for $0<s<d$ 
where $C$ is a constant depending on $d$ and $s$~\citep{frank2025minimizers}. 
Using the Parseval's theorem, \eqref{eq:flow_property} translates to the following inequality in the complex (Fourier) domain 
\begin{align}
     C\int \frac{|\widehat{\mu}(w)|^2}{\|w\|^{2}} dw = \int \frac{1}{\|y-x\|^{d-2}}d\mu_t(x)d\mu_t(y) \leq (d-2) E(\mu_0) \label{eq:fourier} 
\end{align}
holds. Similarly, we can write \eqref{eq:Lcondition} as
\begin{align}
    \int \frac{1}{\|w\|} |\widehat{\mu}(w)| dw \lesssim (1+E(\mu_0))^2 \label{eq:fourer_desired} 
\end{align}

Define set $A = \{ w|\|w\|<1\}$, a straightforward application of Cauchy-Schwarz obtains
\begin{align}
    \left(\int_{A} \|w\|^{-1} |\widehat{\mu}(w)| dw\right)^2 & \leq  \int_A \|w\|^{-2} |\widehat{\mu}(w)|^2 dw \int_A dw \\ 
    & \stackrel{\eqref{eq:fourier}}{\lesssim} E(\mu_0)  
\end{align}
To establish \eqref{eq:fourer_desired}, we need to bound the above integral taken over the complement of $A$ denoted by $A^c$:   
\begin{align}
 \int_{A^c} \frac{1}{\|w\|} |\widehat{\mu}(w)|dw  \leq \int |\widehat{\mu}(w)|dw \stackrel{\eqref{eq:L1boundfourier}}{\lesssim} (1+E(\mu_0))^2
\end{align}
Combining the last two inequality concludes \eqref{eq:fourer_desired}, hence the proof is complete. 


\end{proof}

\subsection{Proof of the Auxiliary Lemma~\ref{lemma:backward}} \label{app:proof_deri_conv}
\begin{repeatlemma}{lemma:backward}
    Let $y_0$ be a random vector drawn from $\mu_0$ such that $E(\mu_0)<\infty$. Consider the following ODE
    \[
    \frac{dy}{dt} = - F_t(y_t), \quad F_t(y) \coloneq \int \nabla W(y - x) \mu_t(x) dx dt
    \] 
    Then, the  distribution of $y_t$ is $\mu_t$. 
\end{repeatlemma} 
\begin{proof} 
    If the vector field $F_t$ obeys 3 conditions in Lemma~\ref{lemma:transport}, then invoking the lemma concludes the proof. It remains to to validate necessary conditions for the vector field. Since $\mu_t$ is a gradient flow, it obeys 
    \begin{align}
        \int \frac{1}{(d-2)\|x-y\|^{d-2}}d\mu_t(x)d\mu_t(y) + \frac{1}{2} \int \| x -y \|^2 d\mu_t(x) d\mu_t(y) = E(\mu_t) \leq E(\mu_t) \label{eq:objective}
    \end{align}
    \paragraph{Condition 1: $\int \| F_t(x)\|\mu_t(x) dx < \infty$.}
    According to the definition, we have 
    \begin{align}
        \| F_t(x)\|\mu_t(x) dx & = \| \int \frac{x-y}{\|x-y\|^d} + (x-y) d\mu_t(y)\|\mu_t(x) dx \\
        & \leq \int \frac{1}{\|x-y\|^{d-1}}d\mu_t(x)d\mu_t(y) + \int \|x-y\|d\mu_t(x) d\mu_t(y)   
    \end{align}
    We bound each term in the above upper-bound. The first term can be bounded  
    using Jensen's inequality as
    \begin{align}
        \int \| x-y \| d\mu_t(x) d\mu_t(y) \leq \sqrt{\int \|x-y\|^2 d\mu_t(x) d\mu_t(y)} \stackrel{\eqref{eq:objective}}{\leq} \sqrt{E(\mu_0)} 
    \end{align}
    Similarly, we get 
    \begin{align}
     \int \frac{1}{\|x-y\|}d\mu_t(x) d\mu_t(y) \leq \left(\int \|x-y\|^2 d\mu_t(x) d\mu_t(y)\right)^{1/(d-2)}\stackrel{\eqref{eq:objective}}{\leq} (E(\mu_0))^{1/(d-2)}
    \end{align}
    The above inequality yields 
    \begin{align}
        \int \frac{1}{\|x-y\|^{d-1}} d\mu_t(x) d\mu_t(y) & \leq \left(\int \frac{1}{\|x-y\|^{d-2}} d\mu_t(x) d\mu_t(y) \right)  \int \frac{1}{\|x-y\|} d\mu_t(x) d\mu_t(y) \\
        & \leq E(\mu_0)^{(d-1)/(d-2)}
    \end{align}

\paragraph{Condition 2. $\| F_t(x)\|< \infty$ for a bounded $x$.} Since all statements hold up to translation, we can assume $\int y \mu_t(y) dy = 0$. This allows us to simplify the expression for the vector field as 
\begin{align}
    \| F_t(x)\| &= \| \frac{x-y}{\|x-y\|^{d}} d\mu_t(y) + x - \underbrace{\int y \mu_t(y)}_{=0}  \| \\ 
    & \leq \int \frac{1}{\|x-y\|^{d-1}}d\mu_t(y)+ \|x\| \\ 
    & \stackrel{\eqref{eq:Lcondition}}{\leq} c_1 +c_2 (1+E(\mu_0)^2) + \|x\|,
\end{align}
where $c_1$ and $c_2$ are constants independent from $t$. 
\paragraph{Condition 3. Lipschitzness.} Lemma~\ref{lemma:Lipschitzness} ensures $F_t$ is Lipschitz.
\end{proof}

\newpage
\section{Proof of \Cref{prop:proximal}}

\begin{repeatprop}{prop:proximal}
    Consider the following proximal optimization problem 
    \begin{align} 
    (y^*_1, y^*_2, \cdots, y^*_n) = \arg\min_{y_1, y_2, \cdots, y_n \in \mathbb{R}^d} \frac{1}{2} \sum_{i = 1}^d \| y_i -x_i^{(k)} \|^2 - \gamma E^{(\epsilon)}_{n}(y_1, y_2, \cdots, y_{n}) \numberthis{eq:prox_inverse}
    \end{align}
    The above optimization is convex with solution $y_i^* = x_i^{(k-1)}$ for all $i \in [n]$, as long as the learning rate $\gamma$ is sufficiently small.
\end{repeatprop}
\begin{proof}
    Let the objective of the proximal step~(\ref{eq:prox_inverse}) defined as,
\[
H (y_1, \cdots, y_n) \coloneqq \frac{1}{2} \sum_{i = 1}^n \| y_i - x_i^{(k)} \|^2 - \gamma E^{(\epsilon)}_n(y_1, \cdots, y_n). \numberthis{eq:objective_proximal}
\]
And for ease of notation, let $Y = (y_1, \cdots, y_n) \in \mleft(\mathbb{R}^{d}\mright)^n$. It is easy to observe that,
\[
\nabla^2 H (Y) = I_{n \times d} - \gamma \nabla^2 E^{(\epsilon)}_n (Y).
\]
Thus, the optimization problem~(\ref{prop:proximal}) is convex as long as we can prove an upper bound on the spectral norm of $\nabla^2 E^{(\epsilon)}_n (Y)$ and choose a small learning rate for the inverse gradient map $\gamma$. From this point onward, we drop the subscript/superscript \( \epsilon \) to streamline the notation.
Thus $W$ is defined as
\[
W(z) = - \frac{m}{(\| z \|^2 + \epsilon)^{m/2}} + \frac{1}{2} \| z \|^2
\]
with direct calculations we have,
\[
\nabla W(z) = \mleft( 1 + m^2 (r^2 + \epsilon)^{-\frac{m}{2} - 1} \mright) z,
\]
where $r \coloneqq \| z \| $. Again with differentiation,
\[
\nabla^2 W(z) = \mleft( 1 + m^2 (r^2 + \epsilon)^{-\frac{m}{2} - 1} \mright) I_d - m^2 (m + 2) (r^2 + \epsilon)^{-\frac{m}{2} - 2} z z^\top.
\]
Hence, for every unit vector $\nu \in \mathbb{S}^d$,
\[
\nu^\top \nabla^2 W(z) \nu & \leq  1 + m^2 (r^2 + \epsilon)^{-\frac{m}{2} - 1} \\
& \leq 1 + m^2 \epsilon^{-\frac{m}{2} - 1}.
\numberthis{eq:spectral_particle}
\]

Recall that,
\[
E_n(Y) = \frac{1}{n(n-1)} \sum_{i=1}^n \sum_{j \neq i} W(y_i - y_j).
\]
Taking derivatives, for each $i$ we have,
\[
\nabla_{y_i} E_n (Y) = \frac{1}{n(n-1)} \sum_{j \neq i} \nabla W(y_i - y_j).
\]
Next, for $j\neq i$,
\[
\nabla^2_{y_i, y_j} E_n(Y) = \frac{- 1}{n(n-1)} \nabla^2 W(y_i - y_j)
\]
and,
\[
\nabla^2_{y_i} E_n(Y) = \frac{1}{n(n-1)} \sum_{j \neq i} \nabla^2 W(y_i - y_j).
\]
Taking an arbitrary block vector $\nu \in \mleft(\mathbb{R}^d\mright)^n$,
\[
\langle \nu, \nabla^2 E_n(Y) \nu \rangle & = \frac{1}{n(n-1)} \mleft( \sum_{i = 1}^n \nu_i^\top \left( \sum_{j \neq i} \nabla^2 W(y_i - y_j) \right) \nu_i + \sum_{i \neq j} \nu_i^\top \left(- \nabla^2 W(y_i - y_j)\right) \nu_i \mright) \\
& = \frac{1}{n(n-1)} \mleft( \frac{1}{2} \sum_{i \neq j} (\nu_i - \nu_j)^\top \nabla^2 W(y_i - y_j) (\nu_i - \nu_j) \mright).
\]

In turn, by spectral bound on each block in~(\ref{eq:spectral_particle}),
\[
\langle \nu, \nabla^2 E_n(Y) \nu \rangle & \leq \frac{1}{2 n(n-1)} \mleft( \sum_{i \neq j} \mleft(\mleft( 1 + m^2 (\|y_i - y_j \|^2 + \epsilon)^{-\frac{m}{2} - 1} \mright) \mright) \| \nu_i - \nu_j \|^2  \mright) \\ \numberthis{eq:spectral_temp1}
& \leq \frac{1}{2 n(n-1)} \mleft( \sum_{i \neq j} \mleft(1 + m^2 \epsilon^{-\frac{m}{2} - 1} \mright) \| \nu_i - \nu_j \|^2  \mright)  \\
& = \frac{\mleft(1 + m^2 \epsilon^{-\frac{m}{2} - 1}\mright)}{2 n(n-1)} \mleft( \sum_{i \neq j} \| \nu_i - \nu_j \|^2  \mright) \\
& \leq \frac{\mleft(1 + m^2 \epsilon^{-\frac{m}{2} - 1}\mright)}{2 n(n-1)} \mleft( 2 n \sum_{i} \| \nu_i\|^2  \mright) \\
& = \frac{1 + m^2 \epsilon^{-\frac{m}{2} - 1}}{n - 1} \| \nu \|^2,
\]
where in (\ref{eq:spectral_temp1}) we used the (\ref{eq:spectral_particle}.

Hence, for the operator norm of $\nabla^2 E_n(Y)$ we infer that,
\[
\| \nabla^2 E_n(Y) \|_2 \leq \frac{1 + m^2 \epsilon^{-\frac{m}{2} - 1}}{n - 1}.
\]
Thus, by choosing any sufficiently small $\gamma \in (0, \frac{n - 1}{1 + m^2 \epsilon^{-\frac{m}{2} - 1}})$, 
\[
\nabla^2 H (Y) \succ 0,
\]
and objective of the proximal step~(\ref{eq:objective_proximal}) is strongly convex and hence is uniquely minimized.

Considering the optimal solution $(y^*_1 ,y^*_2, \cdots, y^*_n)$ to (\ref{eq:objective_proximal}), by first order optimality condition,
\[
\nabla_{y_i} H(y^*_1 ,y^*_2, \cdots, y^*_n) = y_i^* - x_i^{(k)} - \gamma \nabla_{y_i} E_n (y^*_1 ,y^*_2, \cdots, y^*_n) = 0.
\]
Thus,
\[
x_i^{(k)} = y_i^* - \gamma \nabla_{y_i} E_n (y^*_1 ,y^*_2, \cdots, y^*_n),
\]
and hence for all $i \in [n]$,
\[
y_i^* = x_i^{(k-1)},
\]
and proof is concluded.
\end{proof}

\newpage
\section{Experiments}
\subsection{Autoencoder details}
As previously mentioned, the experiments in Section~\ref{sec:experiments} employ an autoencoder. In this section, we provide further details on the architecture and training procedure of the autoencoder. The model is a standard convolutional autoencoder consisting of two encoding layers followed by two decoding layers. The structure of the encoder is summarized in Listing~\ref{auto}. Importantly, the encoder serves as a dimensionality reduction mechanism. Without this reduction, computing \( W^{(s)} \), as required in EFS, becomes numerically infeasible.

\lstdefinestyle{mypython}{
    language=Python,
    backgroundcolor=\color{gray!10},
    basicstyle=\ttfamily\small,
    keywordstyle=\color{blue},
    stringstyle=\color{orange},
    commentstyle=\color{green!50!black},
    showstringspaces=false,
    frame=single,
    breaklines=true,
    captionpos=b
}

\subsection{Training Protocol}
We trained the autoencoder on the MNIST dataset using the Adam optimizer with a learning rate of \(10^{-3}\), a batch size of 250, and for 120 epochs. The training loss, measured by mean squared error (MSE), decreased to 0.008 after 120 epochs. Training was conducted on an NVIDIA RTX 6000 GPU with 48\,GB of memory. The implementation was done in PyTorch~\citep{paszke2019pytorch} (see the notebook \texttt{neurips-figures4-5.ipynb} for details). No preprocessing was applied to the data.
\begin{lstlisting}[style=mypython, caption={Autoencoder},label=auto]
Autoencoder(
  (encoder): Sequential(
    (0): Conv2d(1, 16, kernel_size=(3, 3), stride=(2, 2), padding=(1, 1))
    (1): ReLU(inplace=True)
    (2): Conv2d(16, 32, kernel_size=(3, 3), stride=(2, 2), padding=(1, 1))
    (3): ReLU(inplace=True)
    (4): Conv2d(32, 15, kernel_size=(7, 7), stride=(1, 1))
  )
  (decoder): Sequential(
    (0): ConvTranspose2d(15, 32, kernel_size=(7, 7), stride=(1, 1))
    (1): ReLU(inplace=True)
    (2): ConvTranspose2d(32, 16, kernel_size=(3, 3), stride=(2, 2), padding=(1, 1), output_padding=(1, 1))
    (3): ReLU(inplace=True)
    (4): ConvTranspose2d(16, 1, kernel_size=(3, 3), stride=(2, 2), padding=(1, 1), output_padding=(1, 1))
    (5): Sigmoid()
  )
)
\end{lstlisting}


\end{document}